\documentclass[10pt,twocolumn,letterpaper]{article}

\usepackage{cvpr}              % To produce the CAMERA-READY version
\usepackage[dvipsnames]{xcolor}

\definecolor{cvprblue}{rgb}{0.21,0.49,0.74}
\usepackage[pagebackref,breaklinks,colorlinks,citecolor=cvprblue]{hyperref}

\usepackage{times}
\usepackage{epsfig}
\usepackage{graphicx}
\usepackage{amsmath}
\usepackage{amssymb}
\usepackage[ruled]{algorithm2e}
\usepackage{multirow}
\usepackage{makecell}
\usepackage{booktabs}
\usepackage{amssymb}
\usepackage{pifont}

\usepackage[font=footnotesize]{caption}

\usepackage{paralist}

\usepackage[dvipsnames]{xcolor}

\usepackage[pagebackref,breaklinks,colorlinks]{hyperref}

\usepackage[capitalize]{cleveref}
\crefname{section}{Sec.}{Secs.}
\Crefname{section}{Section}{Sections}
\Crefname{table}{Table}{Tables}
\crefname{table}{Tab.}{Tabs.}

\newcommand{\NOTE}[1]{\textcolor{red}{[NOTE: #1]}}
\newcommand{\abc}{}%[1]{\textcolor{black}{#1}}
\newcommand{\abcn}{}%[1]{\textcolor{magenta}{#1}}
\newcommand{\abcnn}{}%[1]{\textcolor{blue}{#1}}

\newcommand{\jimmy}{}%[1]{\textcolor{cyan}{#1}}
\newcommand{\jimmyy}{}%[1]{\textcolor{violet}{#1}}

\newcommand{\calI}{{\cal I}}
\newcommand{\bz}{\mathbf{z}}
\newcommand{\bx}{\mathbf{x}}
\newcommand{\by}{\mathbf{y}}
\newcommand{\real}{\mathbb{R}}
\newcommand{\bg}{\mathbf{g}}
\newcommand{\bm}{\mathbf{m}}
\newcommand{\bbf}{\mathbf{f}}
\newcommand{\calX}{{\cal X}}
\newcommand{\bI}{\mathbf{I}}
\newcommand{\bh}{\mathbf{h}}
\newcommand{\bV}{\mathbf{V}}
\newcommand{\calF}{{\cal F}}
\newcommand{\calD}{{\cal D}}
\newcommand{\calE}{{\cal E}}
\newcommand{\calP}{{\cal P}}
\newcommand{\calS}{{\cal S}}
\newcommand{\calB}{{\cal B}}
\newcommand{\calA}{{\cal A}}
\newcommand{\calN}{{\cal N}}

\newcommand{\bepsilon}{\boldsymbol{\epsilon}}
\newcommand{\bzero}{\mathbf{0}}
\newcommand{\EV}{\mathbb{E}}
\newcommand{\bbw}{\mathbf{w}}

\newcommand{\CUT}[1]{}

\def\paperID{8} % *** Enter the Paper ID here
\def\confName{CVPR}
\def\confYear{2024}

\title{Learning Tracking Representations from Single Point Annotations}

\author{Qiangqiang Wu  \hspace{30mm} Antoni B. Chan\\
Department of Computer Science, City University of Hong Kong\\
{\tt\small qiangqwu2-c@my.cityu.edu.hk, abchan@cityu.edu.hk}
}

\begin{document}
\maketitle

\begin{abstract}

% existing bounding box annotation expansive....
% propose to learn representations via single point annotaion;
% Specifically, this new schema only requires annotatowrs to click ...., which is 4X faster than xx.
% To learn xx, we propose a soft contrastive learning framework, which use a soft generated positive xx ;
% We show that the proposed soft contrastive learning framework is generic, which can be applied to various frameworks, including the pure Siamese matching framework, correlation filter framework and scale regression-based framework.
% Compared with the supervised baseline, the experimental results show that our method can achieve > 95% performance by only using 22% annotation cost. Further more, w/ the same annotation cost, our method is better than the bounding box annotation;

%This paper tackles the problem of learning deep representations from single point annotations 

Existing deep trackers are typically trained with large-scale video frames with annotated bounding boxes. However, these bounding boxes are expensive and time-consuming to annotate, \abc{in particular for large scale datasets}. 
In this paper, we propose to learn tracking representations from single point annotations (i.e., $4.5\times$ faster to annotate than the traditional bounding box) in a weakly supervised manner. Specifically, we propose a soft contrastive learning (SoCL) framework that incorporates target objectness prior into end-to-end contrastive learning. Our SoCL consists of adaptive positive and negative sample generation, which is memory-efficient and effective for learning tracking representations. We apply the learned representation of SoCL to visual tracking and show that our method can 1) achieve better performance than the fully supervised baseline trained with box annotations under the same annotation time cost; 2) achieve comparable performance of the fully supervised baseline \jimmy{by using the same number of training frames} and \jimmy{meanwhile reducing annotation time cost by 78\% \abcnn{and total fees by 85\%}}; 3) be robust to annotation noise.
\end{abstract}

\vspace{-0.5cm}
\section{Introduction}
\label{sec:intro}
Visual object tracking is a basic computer vision task with a long history spanning decades. In recent years, considerable progress \cite{siamrpn,transt,pul} has been made in the tracking community with the development of deep learning techniques. Deep trackers have achieved strong performance on existing tracking benchmarks \cite{got10k,lasot,trackingnet}, and show great potential in various applications. %, such as robotics and surveillance.

% learning-based trackers \cite{SiamFC,SiamRPN_plus} (deep trackers) mainly benefit from  large-scale annotated tracking datasets for representation learning. These data-driven

%\definecolor{amber}{rgb}{1.0, 0.75, 0.0}
%\definecolor{aureolin}{rgb}{0.99, 0.93, 0.0}
  \begin{figure}
\begin{center}
   \includegraphics[width=0.9\linewidth]{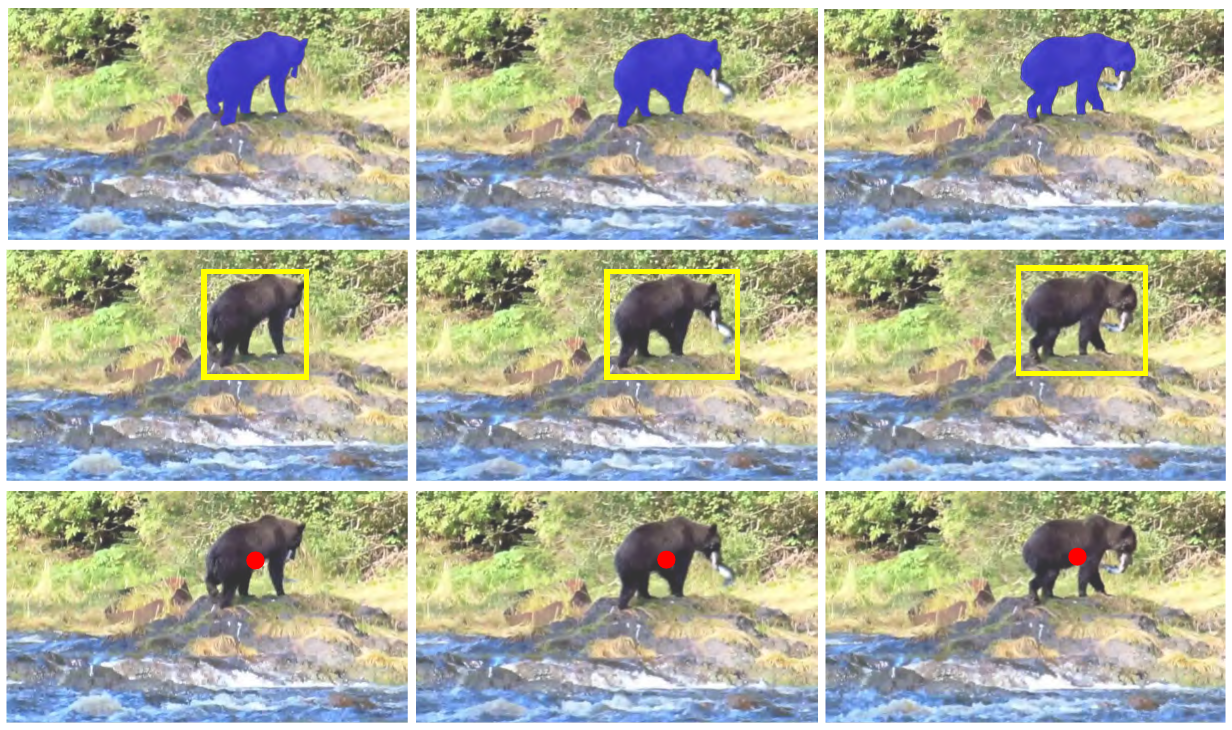} %overall6.eps
\end{center}
\vspace{-0.5cm}
 \caption{An illustration of video frame annotations using masks,
 % (\NOTE{to plot} \jimmy{A: Fixed it.}), 
 bounding boxes and center points. \abc{The time for humans to label} point annotations is $4.5\times$ and $34.4\times$ faster than the time for bounding boxes and mask annotations, respectively. In this paper, we propose a novel {\em soft contrastive learning framework} to learn tracking representations from point annotations in video frames \abc{so as to reduce annotation  cost and \abcnn{total fees}}.}
 \vspace{-0.5cm}
\label{annotation}
\end{figure}

Existing deep trackers are mainly trained with large-scale datasets comprising bounding box annotations on video frames. 
%In practice,
 In order to obtain high-quality bounding box annotations, one common practice is to employ large numbers of people on %some popular 
 a crowd-sourcing platform (e.g., Amazon Mechanical Turk) for annotating. Usually, there are two typical steps to annotate a video frame: 1) draw a  bounding box that tightly includes the object, and 2) verify the annotated bounding box for quality control.  These two steps respectively take 10.2 and 5.7 seconds \cite{boxanno}. Considering that existing tracking datasets consist of millions of annotated bounding boxes, e.g., ILSVRC \cite{ILSVRC} (2.5M) and Got-10K \cite{got10k} (1.4M), a conservative estimate of the time cost for annotating ILSVRC and Got-10K are 7,083 and 3,967 hours, even without accounting for  the verification time. To ease the annotation cost in visual tracking, recent progress \cite{pul,ludt} on unsupervised tracking generate pseudo labels for representation learning. However, these works still lag behind their fully supervised  counterparts \cite{SiamRPN_plus,transt,siamdw} due to noise in the pseudo labels.

 \begin{figure*}
\begin{center}
   \includegraphics[width=0.8\linewidth]{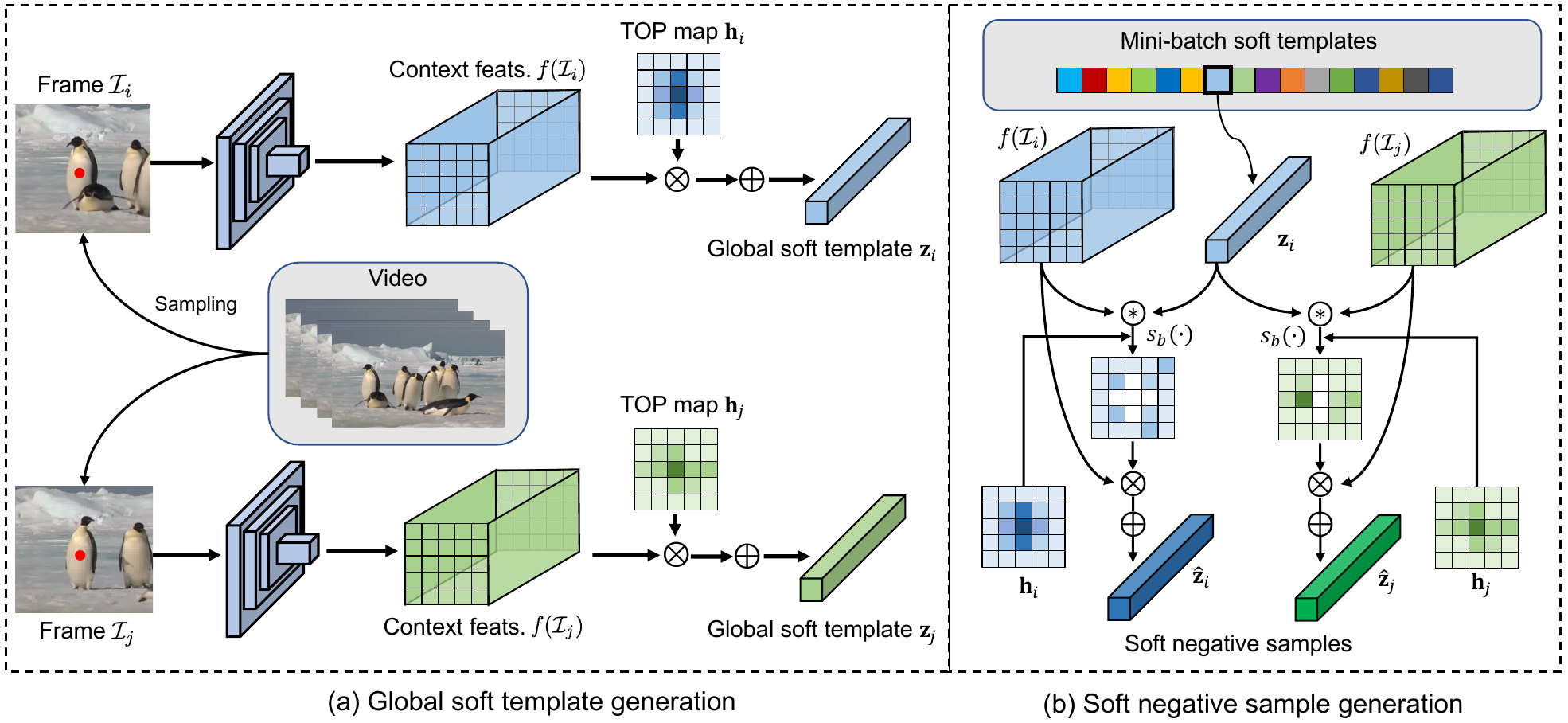} %overall6.eps
\end{center}
\vspace{-0.5cm}
 \caption{Overview of (a) global soft template generation (GST);  and (b) soft negative sample (SNS) generation in the proposed SoCL framework.
  \jimmy{(a) Given two randomly selected frames $\calI_{i}$ and $\calI_{j}$ in a video, we firstly extract their context features $f(\calI_{i})$ and $f(\calI_{j})$, and then calculate GSTs $\bz_i$ and $\bz_j$ as the weighted sum over the spatial locations on $f(\calI_{i})$ and $f(\calI_{j})$, where each location weight is from the corresponding location in the target objectness prior (TOP) maps $\bh_i$ and $\bh_j$. (b) During the mini-batch training, for a specific GST (e.g., $\bz_{i}$), we obtain two similarity maps between $\bz_{i}$ and each location in the context features by using a cross-correlation operation (denoted as $\circledast$). We next use a background selection function $s_{b}(\cdot)$ to mask out target responses and select background counterparts with high responses in the similarity maps to generate the SNSs $\hat{\bz}_i$ and $\hat{\bz}_j$. The generation of both GST and SNS is memory-efficient.}
%  \NOTE{need to fix the caption below...}
%\NOTE{need to put a "summation" node after the circle-x for computing the soft template.}
%\NOTE{In (b), show that the center parts of the weight map are zeroed out due to the object map. Now it is unclear why the middle is missing.}
%\NOTE{in (b), the first set of circle-x should be convolution (circle-*)?}
\abcn{$\otimes$ is element-wise multiplication,  while $\oplus$ is a sum over spatial locations.}
%\NOTE{need to change $I_i$,$I_j$ to $\calI_i$,$\calI_j$ in the figure. \jimmyy{A: fixed it.}}
%\NOTE{add $\hat{\bz}_i$ and $\hat{\bz}_j$ for the output SNS in (b).}
%\NOTE{change "Norm. obj. map" to "TOP map", $\sigma(\bh)$ to $\bh$}
 }
 \vspace{-0.5cm}
\label{socl}
\end{figure*}

Different from previous trackers that use expensive bounding box annotations for fully supervised training, in this paper, we propose to learn tracking representations from low-cost and efficient single point annotations (see Fig.~\ref{annotation}) in a weakly supervised manner. %Given a video frame, our framework enables the 
For point annotations, annotators only need to click once at the object center, which takes about 2.27 seconds per frame (refer to Sec.~\ref{datapre} for more details) and is $4.5\times$ faster than bounding box annotation.  Although point annotations have low time cost, learning effective tracking representations from them is challenging due to the following two reasons: 1) the point annotations naturally lack target scale information, whereas target scale is vital information needed for training traditional deep trackers; 2) the annotations may be noisy since the annotated target center does not always perfectly match the ground-truth target center.

To tackle the above problems and learn robust tracking representations from point annotations, we propose a soft contrastive learning framework (SoCL) that generates global and local soft templates (GSTs and LSTs)
%(i.e., at the feature level) 
based on a target objectness prior (TOP) map, and then optimizes a pairwise contrastive loss between positive/negative soft samples. 
%\NOTE{insert a sentence(s) about the target objectness prior. }
%\NOTE{insert one sentence about how the global soft template is generated.}
\jimmy{The TOP map contains the pixel-wise probabilities that each pixel location belongs to the target. The GSTs are generated by aggregating each location in the  feature map based on the TOP maps (see Fig.~\ref{socl}).}
%Note that  although proposal generation methods are well applied in visual tracking, we are the first to incorporate target objectness prior into end-to-end contrastive learning for visual tracking. 
%Given a global soft template, 
In order to facilitate discriminative feature learning and avoid large memory cost, we propose a memory-efficient method to adaptively generate soft negative samples using high-similarity regions of the cross-correlation map between the GST %global soft template 
and the feature map.
% which is calculated by applying the global soft template to perform the cross-correlation on the context feature map.
%The obtained cross-correlation map highlights the  local regions with high similarities to the soft template, and these local regions are assigned with more weights during the generation of soft negative samples. 
In addition, we also sample \jimmy{LSTs, \abcnn{which simulate partial occlusion or appearance variations}, to augment the positive set, further boosting the representation learning.}
%\NOTE{are the "local positive views" the same as the "local soft template"? \jimmy{Yes, the same. I use the local soft templates instead.}}

The learned representations of SoCL can be directly applied to both Siamese and correlation filter tracking frameworks. 
%The experimental results on several popular tracking benchmarks demonstrate that our method can 1) achieve comparable performance of the fully supervised baseline trained with box annotations by reducing 78\% annotation cost; 2) obtain better performance by using the same annotation cost and 3) be robust to annotation noise.
In addition, we also successfully combine our framework with additional sparse bounding box annotations so as to generate pseudo bounding box labels, in order to train state-of-the-art scale regression-based trackers (e.g., TransT \cite{transt}). %\NOTE{And the representations learned by SoCL are also shown to be effective for the other tracking tasks including multiple object tracking, video object segmentation and pose tracking.\NOTE{remove later?} \jimmyy{A: yes, I will remove it in the main paper.}}
In summary, our main contributions are:
\begin{compactitem}
  \item 
      We propose a soft contrastive learning (SoCL) framework, which incorporates a target objectiveness prior into end-to-end 	  contrastive learning, in order to learn tracking representations from single point annotations. 
  \item
     We propose a memory-efficient method for soft negative sample generation, which %adaptively generates soft negative samples for each global soft template and
      significantly increases the number of negative samples with low memory cost.
    \item
     We propose to  generate \jimmy{a local soft template} for each global soft template and facilitate the representation learning via global-to-local contrastive learning.
  \item 
      Experiment results show that our tracker learned from single point annotations  can: 1) achieve comparable performance to the fully supervised baseline trained with box annotations \jimmy{when using the same number of training frames, while reducing the annotation time cost by 78\% \abcnn{and total fees by 85\%}};
      % \NOTE{we reduce annotation cost by 78\%? \jimmy{Yes}};
      2) obtain better performance by using the same annotation time cost, and 3) be robust to annotation noise.
\end{compactitem}

\section{Related Work}
%\NOTE{I didn't edit this section ytet...will wait until it is complete.}
%There are two main types of deep trackers: correlation filter (CF) trackers with deep features, and end-to-end trainable Siamese trackers. 
% are typically inferior to Siamese trackers due to their limited scale estimation ability.
\textbf{Deep Tracking Methods.} Currently, visual tracking is dominated by deep learning-based trackers that are  trained with large-scale annotated datasets. The deep CF trackers \cite{DeepSRDCF,CCOT,ECO,dsnet,10472496,liang_mm18,liang_mm18,sat,9747821} employ deep features for CF tracking. SiamFC \cite{SiamFC} and SINT \cite{SINT} are two pioneering Siamese trackers, which convert visual tracking to a template matching problem. 
Follow-up works aim to more accurately regress target scale
%, many efforts  have been made on learning to accurately regress target scales 
via anchor free \cite{ocean,roam} or anchor based designs \cite{siamrpn,SiamRPN_plus,CRPN}. Although these methods can achieve favorable performance on several tracking benchmarks, they are still inferior to the recent state-of-the-art deep trackers, including online learning-based trackers (e.g., ATOM \cite{atom}, DiMP \cite{dimp}, PrDiMP \cite{prdimp}), and transformer trackers (e.g., TransT \cite{transt}, STARK \cite{spark}, OSTrack \cite{ostrack} and DropTrack \cite{dropmae} ). However, these deep trackers are trained with large-scale tracking datasets with expensive bounding box annotations, which lead to both large annotation time and fee costs. To ease both of the costs, in this paper, we propose to use low-cost point annotations to train the above methods with the proposed SoCL framework.

\textbf{Annotation Types.} Various annotation types have been explored in  computer vision. Bounding box annotations 
%is the most common annotation, which 
have been widely used in various tasks, e.g., object detection \cite{youbb} and object tracking \cite{got10k}. \cite{boxanno}
%, the detailed study  
shows that annotating a bounding box takes $\sim$10.2s. Mask annotations are also used to perform more fine-grained tasks \cite{simvos,SiamMask}, but  takes $\sim$78s per instance \cite{whatspoint}. To ease the annotation time, \cite{clickpoint,whatspoint} propose to learn models from point annotations for object detection and semantic segmentation. However, these methods are designed for static images, which are not applicable for video representation learning. In contrast, our SoCL can effectively learn temporal correspondences from point annotations in videos and be directly applied for online tracking. 

\textbf{Contrastive Learning.} Contrastive learning methods have achieved leading performance on unsupervised representation learning. Commonly, a memory bank is needed to store pre-computed image features for more efficient and effective learning \cite{membank,CPC,CMC,moco,localagg}. Recent works \cite{simclr,simclrextend} propose to use a large batch size (e.g., 8192) to include large numbers of \abcnn{negative} samples in each mini-batch, % contrastive learning 
or even remove the negatives samples to only focus on target prediction \cite{byol,siamcl,abyol}.
%, which can also obtain favorable performance. 
%In addition, except for performing image-level contrastive learning,
In addition, video frame-level contrastive learning is proposed in \cite{framelearning}.  However, these methods are designed to learn from ImageNet \cite{ILSVRC} images or Kinetics videos \cite{kinetic}, which main contain target objects.
%which are well aligned \NOTE{unclear what "well aligned" means here} with targets. 
%
In our case, no explicit target bounding boxes are provided and most of the video frames are noisy (i.e., containing \abcnn{cluttered backgrounds and distractor objects}), which makes existing methods ineffective. To address these issues, we propose to incorporate target objectness prior (TOP) maps into end-to-end contrastive learning from noisy videos.
\indent

\section{Proposed Method}
\label{method}
  Our goal is to learn effective visual tracking representations from low-cost annotations, i.e., point annotations. Moreover, the learned representations should be generic and effective for various deep trackers, including both Siamese \cite{SiamFC,siamdw} and correlation filter \cite{KCF,ECO} trackers.
 An overview of our proposed soft contrastive learning (SoCL) framework is shown in Fig.~\ref{socl}.
\abc{Previous tracking frameworks \cite{SiamFC,KCF,siamrpn,SiamRPN_plus} are trained using bounding box annotations, which contain scale information that defines the extent of the tracked object on the feature map.
 However, in our formulation, point-wise annotations do not provide explicit scale information.
Thus, we first generate a {\em target objectness prior} (TOP) for each image using its point annotation, which estimates the likely extent of the target. \jimmy{To learn more discriminative features, we propose multiple sample generation methods \abcn{based on the TOP map probabilities},  including global soft template generation (GST),  soft negative sample (SNS) generation and local soft template (LST) generation. The generated samples can be used to learn an effective tracking model via our soft contrastive learning loss.}}

\begin{figure}
\begin{center}
   \includegraphics[width=0.8\linewidth]{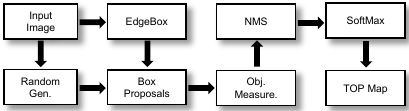} %overall6.eps
\end{center}
\vspace{-0.6cm}
 \caption{Target objectness prior (TOP) map generation for a given input image, which consists of proposal generation (including both EdgeBox and random proposal generation) and aggregation of objectness measurements. 
 %\NOTE{if you have time, you can make a prettier figure, which shows an example for each stage.}
 %\NOTE{add a "softmax" box after "Obj Measure"}
 }
 \vspace{-0.3cm}
\label{pipe_generation}
\end{figure}

\begin{figure}
\begin{center}
   \includegraphics[width=0.8\linewidth]{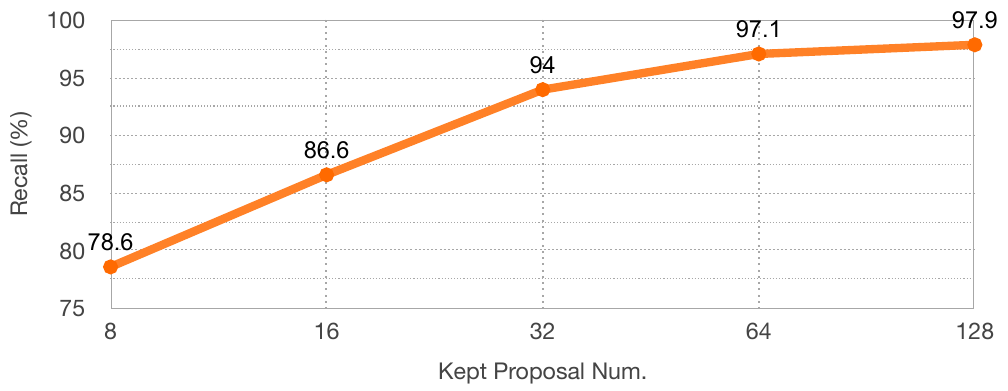} %overall6.eps
\end{center}
\vspace{-0.6cm}
 \caption{A plot of target recall at various numbers of kept proposals after NMS. The evaluation uses an overlap threshold of 0.5.}
 \vspace{-0.3cm}
\label{recall}
\end{figure}

\subsection{\abc{Target Objectness Prior (TOP) Map}}
\label{datapre}

%\NOTE{since we always used the normalized version $\sigma(\bh)$, I changed it so that $\bh$ is the normalized version to reduce the clutter. We can then also call this probabilities then.  Please update the figures ("Norm Obj Map" to "TOP map",  $\sigma(\bh)$ to $\bh$)}

%\NOTE{need a figure to illustrate the process \jimmy{A: fixed it. see Fig. \ref{pipe_generation}}}
%\jimmy{The overall pipeline for generating the target objectness prior (TOP) maps is shown in Fig.~\ref{pipe_generation}}.
\abc{The TOP map contains the probability 
%\NOTE{is it normalized to max value of 1 (so that it is a probability), or should we just call this a score? \jimmyy{A: The original TOP maps are not normalized (i.e., max value is larger than 1. I think we can still call it probability.)}}
 that each pixel belongs to the target object, and is computed by aggregating over the objectness scores of many object proposal boxes sampled over the annotated point (see Fig.~\ref{pipe_generation}).}
%Given a video frame with a point annotation $\calP_{i}$, we aim to generate a target objectness map based on the target center indicated by $\calP_{i}$. 
%For each video frame $\calI_i$ we first generate a target objectness map based on the target center indicated by its point annotation $\calP_i$.
Specifically, given video frame $\calI_i$ and corresponding annotated target center $\calP_i$,  we  generate %multiple (i.e., 
5000 random proposals centered at $\calP_i$ % the annotated target center 
with various scales and aspect ratios. 
Using only proposals centered on $\calP_i$ %, the above pipeline 
may introduce a center bias to the TOP maps, but the point $\calP_i$ might have  spatial annotation noise (since it is difficult to click on the exact center of an object).  Thus, to alleviate this bias, we also generate proposals using EdgeBox \cite{edgebox}. Specifically, for each frame, we use EdgeBox to generate 1,000 proposals and keep the proposals whose center locations are close to the annotated location (i.e., within \jimmy{30 pixels}). These EdgeBox proposals are then combined with the random proposals.

We evaluate the objectness scores for all generated proposals using %an off-the-shell proposal method 
\cite{obj}.
%\NOTE{use the name of the method \jimmy{A: no explicit name for this method, so I continue to use the citation.}}. 
Note that we only use multi-scale saliency, color contrast and edge density cues for the objectness measurement, and exclude the superpixel cue used in \cite{obj}. This variant runs $3.3\times$ faster than the original method and also achieves high recall  in our case (see Fig.~\ref{recall}).
Next, non-maximum suppression (NMS) is applied with an overlap threshold of 0.7 to keep the top-64 proposals for each frame, and filter out redundant proposals.

\begin{figure}
\vspace{-0.1cm}
\begin{center}
   \includegraphics[width=0.8\linewidth]{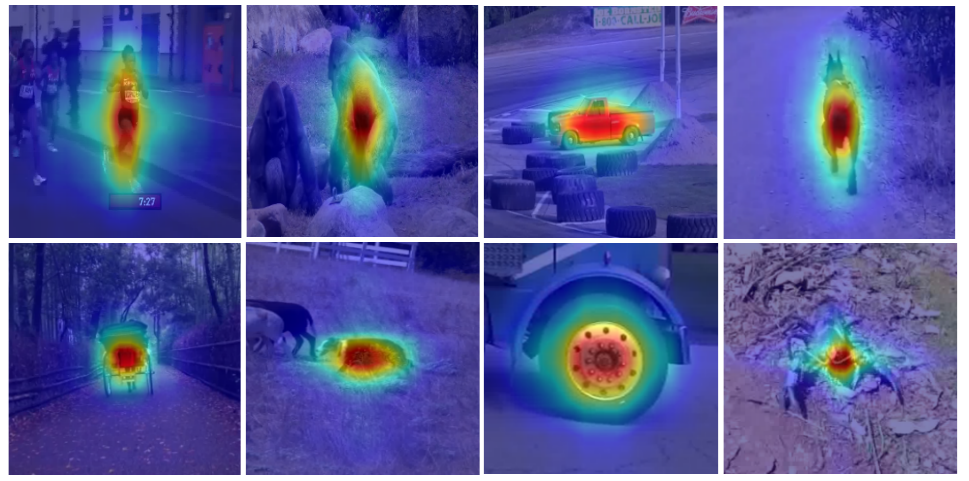} %overall6.eps
\end{center}
\vspace{-0.6cm}
 \caption{Examples of target objectness \jimmy{prior} (TOP) maps generated by using the combination of random and EdgeBox proposal generation.}
\label{target_obj}
\vspace{-0.4cm}
\end{figure}

% with low objectness scores,
Finally, to calculate the target objectness score at each location in the frame, we sum over the scores of all the bounding boxes that cover that location, yielding a score map.
\abcnn{The softmax function (over locations) is then applied to the score map to obtain}
%which yields the %Finally, we obtain 
TOP map $\bh_{i}$ for the video frame $\calI_{i}$. 
Fig.~\ref{target_obj} shows examples of TOP maps. Each score in $\bh$ represents the probability that the corresponding location belongs to the target. Generally, the peak score is located near the annotated location, and the scores gradually decrease moving towards the background regions.

% to the background counterparts.

%These kept proposals are combined with the randomly generated proposals. Following the above pipeline, we use the combined proposals for target objectness map generation. 

In our implementation, calculating the TOP map takes about 0.4 seconds %to efficiently 
for one video frame ($512\times512$), which only needs to be performed once before the training. Since the point annotation in each frame takes about \textbf{1.87} seconds \cite{clickpoint}, the overall per-frame cost of the point annotation and TOP map is \textbf{2.27} seconds, which is \textbf{4.5}$\times$ faster than bounding box annotation (10.2s).

\vspace{-0.1cm}
\subsection{Soft Contrastive Learning}
We next introduce how to use the generated TOP maps for soft contrastive learning.
%\NOTE{add brief summary. define global soft template,  soft negative samples and, local soft template (local positive view?)} 
\jimmy{There are three types of soft samples generated in our SoCL framework, including global and local soft templates (GST and LST) and soft negative samples (SNS), which respectively aggregate global/local target and hard negative counterparts in context features for sample generation.}
% More details are given in the following sections.}

\subsubsection{Global Soft Template (GST) Generation}

% low possibility values are at the positions that are far away from the annotated position.

% in each location of the target objectness map indicates the possibility of the location belonging to the target.

The overall pipeline of global soft template (GST) generation is shown in Fig.~\ref{socl}a. Given a video frame image $\calI_{i}$ and its generated TOP map $\bh_{i}$, the corresponding GST $\bz_{i} \in \real^{C}$ is calculated as a weighted sum over the spatial locations, \abc{with higher weight given to locations that are more likely to be part of the object (according to the TOP map)},
\begin{align}
\label{temp}
\bz_{i} =f(\calI_{i})^{T} \bh_{i},\;  f(\calI_{i}) \in \real^{HW\times C}, \bh_{i}\in \real^{HW \times 1}, 
\end{align}
where %$\sigma(\cdot)$ is the softmax function and 
$f(\cdot)$ is an embedding function (feature extractor), which is implemented as a deep neural network. % in our design. 
The above generation is efficient since it only relies on a single matrix multiplication operation between two $HW$-dim.~vectors. 

During the mini-batch training, for each training video, we randomly select two frames $(\calI_{i},\calI_{j})$ to construct a pair of GSTs $(\bz_{i},\bz_{j})$, which are a pair of positive samples (i.e., the same object). 
%views of each other. 
%Note that $\bz_{i}$ and $\bz_{j}$ are global positive views to each other.
 The GSTs in the other videos are considered as negative samples to $(\bz_{i}, \bz_{j})$. However, these negative samples are usually easy negatives (since that object is completely different), and the number of negative samples are also limited. In the next subsection, we introduce our soft negative sample (SNS) generation for more effective representation learning.

\vspace{-0.3cm}
\subsubsection{Soft Negative Sample (SNS) Generation}
%After obtaining the global soft template $\bz_{i}$ in the frame $\calI_{i}$ of a training video, we sample another frame $\calI_{j}$ in the same video to contrust a temporal corresponding positive pair (i.e., $\bz_{i}$ and $\bz_{j}$) for contrastive learning. 
Previous works \cite{pul,mixup} show that negative samples play an essential role in contrastive representation learning. These methods commonly use a large mini-batch size or specifically design a hard negative selection strategy to include more hard negatives in one mini-batch. However, larger memory cost is also needed for additional negative samples. In this work, we propose to fully leverage the pre-computed context features for memory-efficient soft negative sample (SNS) generation in the feature space. 

In the sampled video, $f(\calI_{i})$ and $f(\calI_{j})$ are regarded as two context feature vectors, which contain both target and background information. The pipeline of SNS generation is shown in Fig.~\ref{socl}b, and the goal is to aggregate hard negative features in the feature map to generate the SNSs $\hat{\bz}_{i},\hat{\bz}_{j} \in \real^{C}$.
%the basic idea is to aggregate hard negative local features to generate soft negative samples $\hat{\bz_{i}} \in \real^{C}$ and $\hat{\bz_{j}} \in \real^{C}$:
\abc{First, the similarity maps $(\bg_{i}, \bg_{j})$ between the GSTs and each location in the feature maps are computed,
\begin{align}
\bg_{i} = f(\calI_{i}) * \bz_{i}, \quad  \bg_{j} = f(\calI_{j}) * \bz_{i},
\end{align}
where $\bg_{i},\bg_{j}  \in \real^{HW}$, and $*$ is the convolution operation. The similarity maps contain high responses for both the target and the hard negative background. 
Next, the high responses for the target are masked out using the TOP maps $(\bh_i, \bh_j)$, which yields the hard-negative maps
\begin{align}
\hat{\bh}_i = \sigma(s_{b}(\bg_{i}, \bh_{i})), \quad \hat{\bh}_j = \sigma(s_{b}(\bg_{j}, \bh_{j})),
\end{align}
where $s_{b}(\cdot)$ is the background selection function (discussed later), and $\sigma(\cdot)$ is the softmax function.
Finally, the SNS $(\hat{\bz}_i, \hat{\bz}_j)$ are generated as a weighted sum of the feature maps over spatial locations, weighted by the hard-negative maps, 
\begin{align}
\hat{\bz}_{i} = f(\calI_{i})^{T} \hat{\bh}_i, \quad \hat{\bz}_{j} = f(\calI_{j})^{T} \hat{\bh}_j, 
\end{align}
Thus the generated SNS contain features from the background with high response to the GST, which helps contrastive learning to find background discriminative features.}

%\begin{align}
%\begin{aligned}
%\bg_{i} &= f(\calI_{i})\bz_{i}, &  \bg_{j} &= f(\calI_{j})\bz_{i},\\
%\hat{\bz_{i}} &= f(\calI_{i})^{T}\sigma[s_{b}(\bg_{i}, \sigma(\bh_{i}))], & \hat{\bz_{j}} &= f(\calI_{j})^{T}\sigma[s_{b}(\bg_{j}, \sigma(\bh_{j}))],
% \end{aligned}
%\end{align}
%where $\bg_{i} \in \real^{HW}$ and $\bg_{j} \in \real^{HW}$ are two generated similarity maps in terms of $\calI_{i}$ and $\calI_{j}$, which indicate the similarity between each location in the context features and the global soft template $\bz_{i}$. $s_{b}(,)$ is a background sampling function, which samples background locations and filters out some target locations (i.e., having large weights in the normalized target objectness maps $\sigma(\bh_{i})$  and  $\sigma(\bh_{j})$) 

%to generate soft negatives. 

% and it guarantees that there is no large mutual information between the generated soft negatives and the soft template.

{\bf Background selection function.}
The background selection function  $s_{b}(\cdot)$ masks out the target object locations in the similarity map $\bg$ using the  TOP map $\bh$.
First, the locations with high score in $\bh$ are selected by thresholding its cumulative sum.  Specifically,
elements in $\bh$ are sorted in descending order, yielding the sorted vector $\bf{a}$ and inverse mapping $\phi(k)$ such that \jimmyy{$h_{\phi(k)} = a_k$}.  Next, the first location that gives a cumulative sum of at least $\theta_b$ is computed,
\begin{align}
\label{neg_sel}
q^* = \min_{\sum_{k=1}^{q} a_k \geq \theta_b} q, 
\end{align}
where $\theta_b \in [0,1]$ is a fixed threshold, and thus the dimensions $\{1,\cdots,q^*\}$ in $\mathbf{a}$ correspond to the target locations with high score.
Second, the corresponding locations in the similarity map $\bg$ are masked out,
% thus yielding 0 weight after applying the softmax,
\begin{align}
s_b(\bg,\bh) = \Big[{\footnotesize \begin{cases}
-\infty, & i \in \{\phi(k)\}_{k=1}^{q^*}, \\
\bg_i, & \mathrm{otherwise}.
\end{cases}}\Big]_i
\label{eqn:sel}
\end{align}
Note that $\theta_{b}$ controls the target location selection, and it should be set to a relatively large value in order to include most of the target locations.  Otherwise, too many target (positive) features are included in the generated SNS, which degrades the learning.
\abcn{In our implementation, $\theta_b=0.8$.}

  \begin{figure}
  \vspace{-0.1cm}
\begin{center}
   \includegraphics[width=0.7\linewidth]{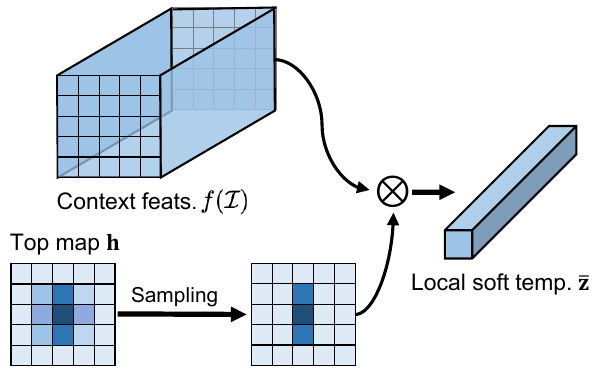} %overall6.eps
\end{center}
\vspace{-0.6cm}
 \caption{\abcn{Overview for generating local soft templates (LST) by sampling high-score locations in the TOP map}
% \NOTE{according to (7) in the text, the sampled map should be either 0 or 1 (here you show 3 values) \jimmyy{A: Yes, the set value should not be 1, I modified Eq. (\ref{posthreshold})}}
 }
 % local positive sample generation. \NOTE{if it's not random, then change  "sampling" into "selection" \jimmy{here we use sampling.}}}
 \vspace{-0.5cm}
\label{local_view_fig}
\end{figure}

 \vspace{-0.2cm}
\subsubsection{Local Soft Template (LST) Generation}
%After obtaining the global soft template $\bz_{i}$ in (\ref{temp}), we then randomly pick another frame $\calI_{j}$ in the same training video and calculate its global soft template $\bz_{j}$. $\bz_{j}$ can be regarded as a global positive view to $\bz_{i}$. 
%\NOTE{can we call this "local soft template (LST)" for consistency with GST? \jimmy{Yes, I change the name to LST.}}

\abc{The TOP map gives the likely extent of the object, but there could be some errors in the map.}
To further enrich positive views and \abc{to make the learning robust to these errors}, we propose to generate a LST $\bar{\bz}$ for each GST $\bz$ by aggregating over a subset of target locations based on the TOP map $\bh$ (see Fig.~\ref{local_view_fig}).
%, the local positive view $\bar{\bz}$ is also guided by the normalized TOP $\sigma(\bh)$ so as to aggregate over a subset of target locations.
\abc{Similar to SNS, first the locations with high score in the  TOP map $\bh$ are selected by finding the top accumulated scores via the selection function
%\jimmyy{
\begin{align}
\label{posthreshold}
s_t(\bh) = \Big[ 
{\footnotesize \begin{cases}
\bh_{i}, & i \in \{\phi(k)\}_{k=1}^{q^*}, \\
0, & \mathrm{otherwise},
\end{cases}}
 \Big]_i, 
\end{align}
where $q^*$ is computed as in (\ref{eqn:sel}) but using threshold $\theta_p$.
Second, the LST is generated as a weighted sum over the feature map, aggregating features with high-probability to be the object,
 \begin{align}
\label{local_view}
\bar{\bz} =f(\calI)^{T}\psi(s_{t}(\bh)),
\end{align}
where $\psi(\cdot)$ is a total sum normalization function.}

\abc{Here $\theta_{p}$ controls how many high-scoring locations are selected for the LST, with $\theta_{p}  \rightarrow 1$ selecting more more complete LSTs ($\theta_{p} =1$ is equivalent to the GST).
}
\abcn{In order to make training robust, we randomly generate the LSTs based on the TOP map probabilities, by sampling $\theta_p$ from a uniform distribution over $[b_p,1)$ each time we generate an LST. 
In our implementation, hyperparameter $b_p=0.6$.
}

 \vspace{-0.2cm}
\subsubsection{Soft Contrastive Learning Loss}

\abc{Each mini-batch in our contrastive learning contains 2 sampled frames from $N$ sampled training videos, totaling $2N$ frames.
The GST, SNS, and LST are generated for each frame in the mini-batch, and then collected to form the positive and negative sample sets for contrastive learning.}

{\bf Negative Sample Set.}
For a GST pair $(\bz_i,\bz_j)$ from the same video, its negative samples come from three sources in the mini-batch:
1) $2(N-1)$ GSTs that are generated from the other $N-1$ videos;
%, where $N$ is the number of sampled videos; 
2) $4N$ SNS generated from all videos; 
3) $2N$ additional hard negative samples created using a mix-up strategy  \cite{mixup}.
Our mix-up strategy generates a novel hard negative example $\hat{\bz}'_i$ for each $\bz_i$ by interpolating its two hardest SNS $(\hat{\bz}_{1}, \hat{\bz}_{2})$, i.e., 
 $\hat{\bz}'_i = \lambda \hat{\bz}_{1} + (1-\lambda) \hat{\bz}_{2}$, where  $\lambda \in (0.5, 1)$ is the interpolation factor.

In total there are $8N-2$ negative samples for each GST pair $(\bz_i,\bz_j)$, which is denoted as the negative sample set 
$\calN_{ij} = \{\hat{\bz}_{k}\}_{k=1}^{8N-2}$.
Our negative sample set is 4$\times$ larger than the baseline of only using the \abcn{positive samples (the $2N-2$ GSTs)} from the other videos. Meanwhile, the SNS generated from  the pre-computed context features have no additional memory cost.

%Finally, we can obtain $(8N-2)$ soft negative samples $\{\hat{\bz_{k}}\}_{k=0}^{8N-2}$ in total for each $\bz$, which is about 4$\times$ larger than the baseline that uses ($2N-2$) negative samples (i.e., global soft templates in the other videos) and meanwhile the soft negatives generated on the pre-computed context features have no additional memory cost, which is memory efficient.

%3) We also employ the negative mix-up strategy \cite{mixup} to further generate novel hard negatives in the feature space. Specifically, for each $\bz$, we use its two hardest negative samples in the above to generate a novel one: $\hat{\bz}^{'} = \lambda \hat{\bz_{1}} + (1-\lambda) \hat{\bz_{2}}$, where $\lambda \in (0.5, 1)$, $\hat{\bz_{1}}$ and $\hat{\bz_{2}}$ are respectively the hardest and 2nd-ranked negative samples to $\bz$. 

\textbf{Global-to-global Contrastive Learning.} Suppose that we treat the global soft template $\bz_{i}$ as a query, based on its global positive view $\bz_{j}$ and the negative sample set $\calN_{ij}$, %the overall negative samples $\{\hat{\bz_{k}}\}_{k=0}^{8N-2}$, 
a global-to-global contrastive learning loss is computed as
% between GSTs $(\bz_{i}, \bz_{j}$) and its negative sample set $\calN_{ij}$:
\begin{align}\label{contras_gg}
\mathcal{L}(\bz_{i}, \bz_{j}, \calN_{ij}) = -\log\tfrac{\exp({\bz_{i}^T \bz_{j}/\tau})}{\sum_{\hat{\bz}_k \in \calN_{ij}} %{k=1}^{8N-2}{
\exp({\bz_{i}^T \hat{\bz}_k/\tau})}, 
\end{align}
where $\tau$ is a temperature hyper-parameter. 

\textbf{Global-to-local Contrastive Learning.} %Except for global-to-global contrastive learning, 
Global-to-local contrastive learning is also conducted to make the learned representations  robust to scale variations and partial occlusion. \abc{The global-to-local contrastive loss uses the LSTs $\bar{\bz}_i$ and $\bar{\bz}_j$ in place of $\bz_j$, which is given by $\mathcal{L}(\bz_{i}, \bar{\bz}_{j}, \calN_{ij})$ and $\mathcal{L}(\bz_{i}, \bar{\bz}_{i}, \calN_{ij})$.}

%\begin{align}\label{contras_gl}
%\mathcal{L}_{g}(\bz_{i}, \bar{\bz}_{j}, \calN_{ij}), 
%\mathcal{L}_{g}(\bz_{i}, \bar{\bz}_{i}, \calN_{ij}) . 
%\end{align}

%\begin{align}\label{contras_gl}
%\mathcal{L}_{g}(\bz_{i}, \bar{\bz_{j}}) = -\text{log}\frac{\exp({\bz_{i} \cdot \bar{\bz_{j}}/\tau})}{\sum_{k=1}^{8N-2}{\exp}({\bz_{i}\cdot \hat{\bz_{k}}/\tau})}, \\
%\mathcal{L}_{g}(\bz_{i}, \bar{\bz_{i}}) = -\text{log}\frac{\exp({\bz_{i} \cdot \bar{\bz_{i}}/\tau})}{\sum_{k=1}^{8N-2}{\exp}({\bz_{i}\cdot \hat{\bz_{k}}/\tau})}, 
%\end{align}

Thus, the overall training loss is for $(\bz_i,\bz_j)$ is the sum of the global-to-global and global-to-local losses:
{\small
\begin{align}\label{overall_loss}
\mathcal{L}_{all}=\mathcal{L}(\bz_{i}, \bz_{j}, \calN_{ij}) + \mathcal{L}(\bz_{i}, \bar{\bz}_{j}, \calN_{ij})  +  \mathcal{L}(\bz_{i}, \bar{\bz}_{i}, \calN_{ij}).
\end{align}
}

%Note that we do not try to set different weighting parameters for the above three loss terms. Treating them equally works well in the end-to-end representation learning.

 \vspace{-0.35cm}
\subsection{Tracking Applications}
\label{ta}

%After learning the model $\mathcal{M}$, we directly integrate it to existing tracking frameworks (e.g., Siamese \cite{SiamFC} and correlation filter (CF) \cite{KCF,ECO} tracking frameworks) as a backbone network without further modifications.

\abc{After training the model with  the overall loss in (\ref{overall_loss}), we use the context feature extractor $f(\calI)$ as the backbone $\mathcal{M}$ for various tracking applications.}

\textbf{Siamese Tracking.} We directly integrate the learned $\mathcal{M}$ into a basic SiamFC \cite{SiamFC} tracking framework without further modifications. The online tracking steps are also same as \cite{SiamFC}, where the template and search features are extracted  using our $\mathcal{M}$ and a cross-correlation operation is applied for target localization. The obtained basic tracker is denoted as ``SoCL-SiamFC'', which can show the effectiveness of our learned backbone $\mathcal{M}$.
%This native SoCL-SiamFC tracker without model updating can be used to effectively show the superiority of the learned $\mathcal{M}$.

%SiamFC \cite{SiamFC} is a typical deep tracker that uses a simple cross-correlation operation for target localization. adopts an Alex-net-like backbone. We implement our backbone $\mathcal{M}$ as the same Alex-net-like backbone and train the model with the proposed soft contrastive learning from scratch. The final learned model is directly integrated to SiamFC without any modifications. The online tracking steps are also same with the original SiamFC tracker.

\textbf{Correlation Filter Tracking.} We also validate our learned backbone $\mathcal{M}$ in an online updating-based correlation filter (CF) tracking framework. CF trackers learn a target appearance model by solving a ridge regression problem in the frequency domain, and the appearance model is continuously updated during online tracking. ImageNet pre-trained models (e.g., VGGNet \cite{vggm}) are usually employed as feature extractors by CF trackers, which can obtain strong performance. Here we employ our learned $\mathcal{M}$ as the feature extractor for the efficient convolution operator-based CF framework \cite{ECO}, and denote it as ``SoCL-CF''. 
%\NOTE{maybe nicer to call it SoCL-ECO. \jimmy{here we use a third-party implementation of ECO, and the implementation is inferior to the original ECO. Also, we do not use Color Names features used in ECO and change the backbone to resnet-18, so I call it SoCL-CF and Res18-CF instead of ECOs}}

%Correlation filter (CF) trackers solve a Redgr regression have an online updating schema to encode target appearance variation online. 	ImageNet Pretrained models are commonly employed as feature extractors. Similar to xx and xx, we use our learned model as a feature exctor in ECO framework \cite{ECO}.  

\textbf{Scale Regression-based Tracking.} State-of-the-art scale regression-based trackers \cite{siamrpn,SiamRPN_plus,transt}  commonly use a bounding box regression branch to estimate the target scale. The training of this regression branch requires bounding box annotations. Here, we propose to combine the point annotations with a set of sparse box annotations, in order to train these scale regression trackers with low annotation cost. Specifically, bounding box annotations are provided in every $T$ frames, such that each training video is divided into multiple short snippets with frame length of $T$. In each snippet, a bounding box defines the tracked object in the first frame and center point annotations are given in the following frames. We then run our SoCL-CF to generate pseudo bounding boxes for each snippet. Note that when the estimated target location by SoCL-CF is far away from the corresponding annotated center point (i.e., $>20$ pixels), we treat it as a tracking failure and use the annotated point as the estimated bounding box center. Meanwhile, the tracker is also corrected to \jimmy{move to the annotated locations}.
%\NOTE{unclear sentence}.
%To more effectively handle large scale variations, s

Our new annotation scheme is also quite efficient and the breakdown of its overall per-frame annotation cost is as follows: 1) overall per-frame point annotation cost \jimmy{$2.27(1-\frac{1}{T})$} (excluding the point annotation for the first frrame);
% \NOTE{actually you don't need the point annotation for the first frame. \jimmy{Yes, now correct it.}}; 
2) running speed of SoCL-CF (0.1s per frame); 3) sparse bounding box annotation cost ($\frac{10.2}{T}$ per frame). We set $T=10$, so the overall per-frame annotation cost for this new scheme is \jimmy{3.16}s, which is about $3.2\times$ faster than dense bounding box annotation. \jimmyy{We use this new schema to generate pseudo bounding boxes on GOT-10k, and then train a scale regression-based tracker, TransT \cite{transt}).} The obtained tracker is denoted as ``SoCL-TransT''. Note that we use the same training configurations in \cite{transt}.
%to train our SoCL-TransT.

 %\vspace{-0.45cm}
\section{Experiments}

\abcn{We present tracking experiments showing the efficacy of our backbone learned from point annotations.
We first present ablation study, and then compare our method versus baselines at different operating points, such using the same annotation cost and using the same training videos.
}

\subsection{Implementation Details}

For SiamFC\cite{SiamFC}, $\mathcal{M}$ is an AlexNet-like network that is randomly initialized, 
while for SoCL-Siam and SoCL-CF,  $\mathcal{M}$  is ResNet-18 \cite{resnet} pre-trained on ImageNet. 
%We implement $\mathcal{M}$ as the randomly initialized Alex-net-like network in SiamFC\cite{SiamFC} and an ImageNet pre-trained ResNet-18 \cite{resnet} network for SoCL-Siam and SoCL-CF. 
For SoCL-CF, we use the features extracted from {\em{Conv1}} and {\em{Layer3}} of ResNet-18.
% for deep correlation filter tracking. 
Different from \cite{ECO}, there are no color names features used in SoCL-CF, and the feature extractor in SoCL-CF\footnote{SoCL-CF is based on a 3rd party implementation of ECO: \url{https://github.com/fengyang95/pyCFTrackers.git}} is ResNet-18 rather than VGG-M \cite{vggm}.
%\NOTE{add a footnote about which 3rd party implemention that you use for ECO \jimmyy{A: Fixed it.}} 
The baseline of SoCL-CF is denoted as Res18-CF, which  uses the ImageNet pre-trained ResNet-18 as feature extractor. We use mini-batches of 192, and Adam optimizer \cite{Adam} with learning rates of 3e-4 for training SoCL-Siam and SoCL-CF. SoCL-Siam and SoCL-CF are trained for 1000 and 500 epochs with a learning rate decay of 0.1 at 500 and 50 epochs, respectively. In addition, $\tau=0.5$, and we experimentally set $\theta_{b}=0.8$ and $b_{p}=0.6$ (see Sec. \ref{ab_study}).

%We implement the  the Alex-net-like backbone network in SiamFC \cite{SiamFC} and the ImageNet pre-trained ResNet-18 \cite{resnet} backbone for SoCL-Siam and SoCL-CF. The backbone in SoCL-Siam is randomly initialized and we train it with our SoCL from scratch. For SoCL-CF, we use the features extracted from {\em{Conv1}} and {\em{Layer3}} of the learned ResNet-18 for deep correlation filter tracking. Different from ECO \cite{ECO}, there is no color names features used in SoCL-CF and the feature extractor in SoCL-CF is ResNet-18 rather than VGG-M \cite{vggm}. We denote the baseline of SoCL-CF as Res 

%We use two backbone networks: the Alex-net-like backbone network in SiamFC \cite{SiamFC}, and a CIResNet-22 backbone network in SiamDW \cite{siamdw}. For fair comparison with the supervised baselines, we use the same network initial parameters as SiamFC and SiamDW, and further train the backbone networks using our proposed PUL. We denote our PUL-based trackers as AlexPUL and ResPUL. 
%\abc{For BD learning, we set temperature $\tau=0.5$ and scale $E=5$.}

\textbf{Point annotations in tracking datasets.}
For fair comparison with box annotations and to better validate the effectiveness of our SoCL , we assume that the point annotation noise is similar to the noise in box annotations. Based on this assumption, we generate center point annotations from the original box annotations in GOT-10k \cite{got10k}, 
%Finally, for each video frame in GOT-10k, we obtain its point annotation, and use all 
%We use these point annotations from GOT-10k to train
% the AlexNet-like and ResNet-18 models in 
and then train
SoCL-Siam and SoCL-CF.

%\noindent\textbf{Noisy point annotations.}

\subsection{Ablation Study}
\label{ab_study}

\begin{figure}
\vspace{-0.3cm}
\begin{center}
   \includegraphics[width=0.7\linewidth]{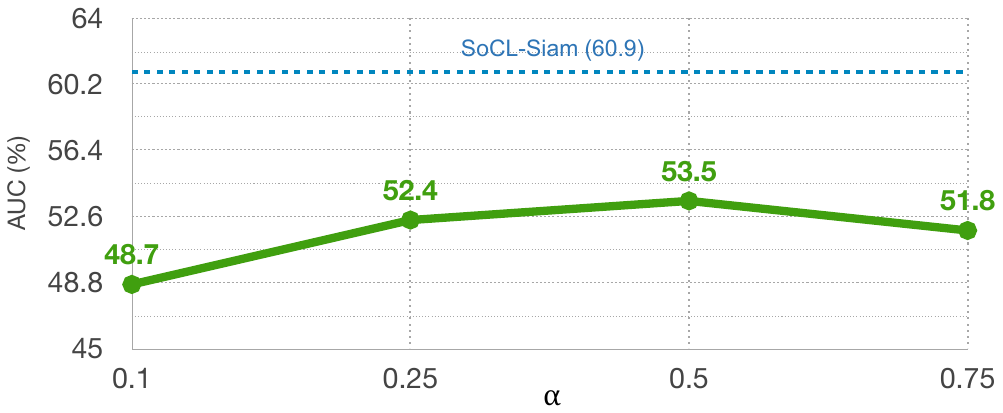} %overall6.eps
\end{center}
\vspace{-0.7cm}
 \caption{Comparison of SoCL-Siam, which is trained with contrastive learning on point annotations, and BCE-Siam, which is trained with BCE loss on pseudo bounding boxes.
 The pseudo bounding boxes are generated from the TOP maps using threshold $\alpha$.
% \NOTE{can save space by shrinking the width of the horizontal axis to 1/4 the page width, and then pairing with Fig 9.}
 }
 \vspace{-0.4cm}
\label{base_box}
\end{figure}

\textbf{Positive/negative sample generation.}
\abcn{In this experiment, we show that there is consistent improvement for each module that generates positive/negative samples for SoCL
% contrastive learning 
(see AUC on OTB-13 and EAO on VOT-16 in Table \ref{ablation}).}
\abcn{First, we only use the GSTs for contrastive learning, where GSTs from one video form the positive set, and those from other videos form the negative set.}
 %generated by our proposed SoCL in contrastive learning 
%can be employed for contrastive learning
 %by using only the global-to-global contrastive loss (\ref{contras_gg}). 
% \NOTE{I removed global-to-global (\ref{contras_gg}) because this also includes the SNS in the denominator.  I assume here we are just using the GSTs for contrastive learning. \jimmyy{A: Yes, that's true. Only GSTs should be used here.}}
 The learned model is integrated to SiamFC for evaluation, and its AUC is % performs
%. As show in Table \ref{ablation}, this baseline performers 
better than the original method learned with a pixel-wise binary-cross entropy (BCE) loss (see Fig. \ref{base_box}), which shows that our SoCL can learn more effective representations from the generated noisy TOP maps than the traditional learning.
%pixel-wise binary-cross entropy learning.

 \newcommand{\cmark}{\ding{51}}%
\newcommand{\xmark}{\ding{55}}%
\begin{table}[t]
  \newcommand{\tabincell}[2]
  \centering
 %\fontsize{9.2}{8}\selectfont   
  \caption{Consistent improvements of AUCs/EAOs (on OTB-13 and VOT-16, respectively) achieved by using global soft template (GST) generation, soft negative sample (SNS) generation, negative mixup and local soft template (LST) generation.}  
   \vspace{-0.3cm}
  \label{ablation}
  \footnotesize
    \centering
    \begin{tabular}{cccccc} 
    \Xhline{\arrayrulewidth}
    \multicolumn{1}{c}{GST Gen.}&  
  \multicolumn{1}{c}{SNS Gen.}& \multicolumn{1}{c}{Neg. Mix.}& \multicolumn{1}{c}{LST Gen.}&\multicolumn{1}{c}{AUC / EAO} \cr
   \Xhline{\arrayrulewidth}       
   %\cmark &                   	&    &         & 		                    			             		&53.5      \cr
    \cmark                  	&    &         & 		                    			             		&55.1 / 0.215      \cr
    \cmark                  	&\cmark    &         & 		                    			                 &57.8 / 0.228   \cr
    \cmark                  	&\cmark    &\cmark         & 		                    			&58.1 / 0.231     \cr
    \cmark                  	&\cmark    &\cmark         &\cmark 		                    		&\textbf{60.9} / \textbf{0.244}   \cr
    
   \Xhline{\arrayrulewidth}  
   \end{tabular}  
      \vspace{-0.2cm}
\end{table}

\begin{figure}
\vspace{-0.2cm}
\begin{center}
   \includegraphics[width=1.0\linewidth]{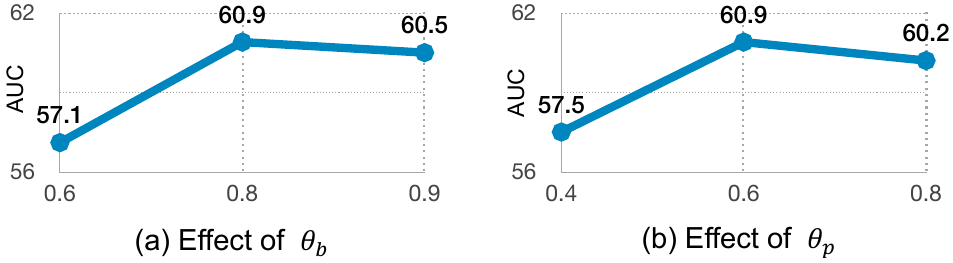} %overall6.eps
\end{center}
\vspace{-0.7cm}
 \caption{Ablation study of $\theta_{b}$ and $\theta_{p}$ that control SNS and LST sample generation respectively on OTB-13.}
 %\NOTE{right figure label should be (b)}
%\NOTE{save vertical space by truncating the lower half of the graphs (which are empty)}}
 \vspace{-0.4cm}
\label{ab_theta}
\end{figure}

Second, 
adding the SNS into the negative set for CL
%\noindent\textbf{Soft negative sample (SNS) generation.} As illustrated in Table \ref{ablation}, the global negative sample generation 
improves AUC by 2.7\% and EAO by 1.3\% -- 
% performance improvement over the baseline on OTB13,
% which is mainly because 
the generated negative samples are hard negative samples that are similar to the positive GSTs, % in the feature space, 
which improves the discrimination power of the tracking model. 
Moreover, the increased number of negative samples also improves the lower bound \jimmyy{(on target mutual information)},
%\NOTE{lower bound of what? \jimmyy{A: lower bound on mutual information}},
 thus leading to better results.

Third, using negative mix-up to further augment the negative set 
%\noindent\textbf{Negative mix-up.} The negative mix-up strategy can 
slightly improves the AUC. Since we perform the negative mix-up in the feature space, there is no additional memory cost, and our SoCL can gain additional benefits. 

Finally, augmenting the positive set using LSTs improves the AUC by 2.8\% to 
%\noindent\textbf{Local soft template (LST) generation.} The positive view augmentation consistently improves the results to
 60.9\% and EAO by 1.3\% to 0.244, %The LSTs significantly improve the model learning. 
 %This is because 
showing that the sampled LSTs effectively mimic targets with partial occlusion or large appearance variations. 
%Learning with LSTs makes the model more robust to various challenges in visual tracking. 

\textbf{Effect of $\theta_{b}$ and $\theta_{p}$.} The ablation study for the effect of $\theta_{b}$ and $\theta_{p}$ is in Fig.~\ref{ab_theta}. Using small values for $\theta_{b}$ degrades the AUC, since the SNSs will contain  too many target features.
Also, using small values of $\theta_p$ also degrades AUC, since the LSTs will not encode enough target features

%or $\theta_{p}$ degrades the performance. This is because the generated SNSs are mixed with too much target counterparts and the LSTs encode too limited target information.

  \begin{figure}[t]
\begin{center}
   \includegraphics[width=0.7\linewidth]{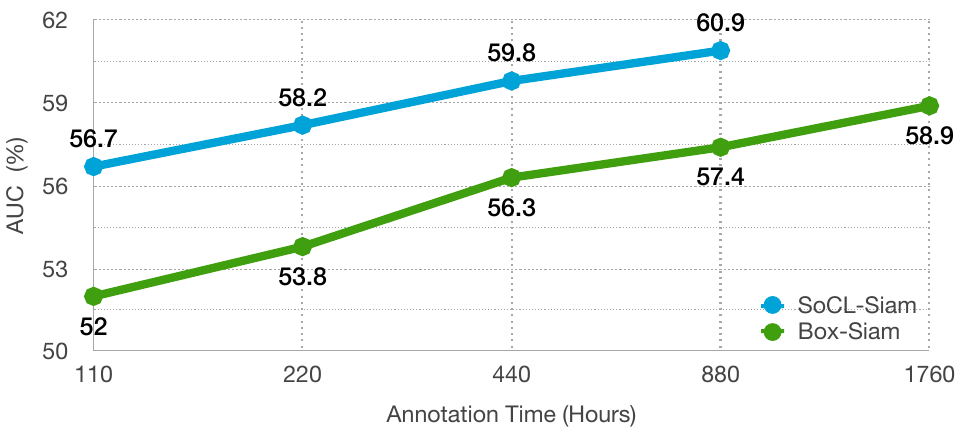} %overall6.eps
\end{center}
\vspace{-0.7cm}
 \caption{Comparison of Box-Siam and SoCL-Siam trained using the same annotation time costs (AUC on OTB-13).}
 \vspace{-0.3cm}
\label{same_anno_otb13}
\end{figure}

\textbf{Learning from pseudo bounding boxes.}  
\abcn{An alternative training framework could use the TOP maps to generate pseudo bounding boxes for standard BCE training.}
%After obtaining the generated TOP maps, instead of conducting our SoCL,
Specifically, we use an adaptive threshold $\alpha$ on each frame to generate a pseudo bounding box, and then %for thresholding in order to generate pseudo bounding boxes. Following 
follow previous methods \cite{SiamFC,siamdw} by training SiamFC with a BCE loss.
The results are shown in  Fig.~\ref{base_box} using different thresholds ($\alpha$).
Using the pseudo bounding boxes and standard BCE loss has degraded performance compared to our SoCL, which is due to the noise in the TOP maps being transferred to the pseudo bounding boxes. In contrast, our soft representations are more robust to the noisy TOP maps.

%severely degrades the performance due to the usage of pixel-wise learning loss (BCE), while our soft representations are more robust to the noisy TOP maps.

\begin{table}[t]
  \newcommand{\tabincell}[2]
  \centering
  \caption{Comparison of Box-Siam and SoCL-Siam trained using the same annotation time costs (i.e., hours) in terms of AUC on OTB-15. The best results are highlighted.}  
   \vspace{-0.2cm}
  \label{same_anno_otb15}
\footnotesize
    \centering
   \begin{tabular}{ccccc}
    \Xhline{\arrayrulewidth}
    \multirow{1}{*}{Annotation time cost}&  
  \multicolumn{1}{c}{110h}& \multicolumn{1}{c}{220h}&\multicolumn{1}{c}{440h}&\multicolumn{1}{c}{880h} \cr
     \Xhline{\arrayrulewidth}  
     Box-Siam              &51.7   &54.0 &55.2 & 57.0      \cr  
     SoCL-Siam &\textbf{53.3}	  &\textbf{54.9} &\textbf{56.5}   &\textbf{57.8}   \cr    
   \Xhline{\arrayrulewidth}  
   \end{tabular}  
   \vspace{-0.45cm}
\end{table}

\vspace{-0.1cm}
\subsection{Comparison with Same Annotation Time Cost}
\abcn{We next compare our SoCL to a fully-supervised baseline trained with bounding-boxes (denoted as Box-Siam) under the same time cost of annotation.}
%%
%To make a fair comparison with the fully supervised baseline (i.e., Box-Siam) trained with bounding box annotations, we compare the proposed SoCL-Siam with Box-Siam under the same annotation time cost. 
%
%We use GOT-10k as the training dataset, and %. Specifically, we 
We randomly sample videos from GOT-10k to meet a specific total annotation time requirement for both Box-Siam and SoCL-Siam. %(e.g., 880 hours)
%
% SoCL-Siam can use more sampled videos with point annotations for training due to the low point annotation time cost. For fair comparison, we make the sampled videos in SoCL-Siam fully overlap the videos used in Box-Siam, such that their training data distribution are similar and the main difference is annotation types.
% \NOTE{the previous two  sentences are contradictory -- for each annotation cost, do SoCL and Box use the same videos, but SoCL uses more frames? \jimmyy{A: No. It seems that this is a better sampling solution for this experiment, i.e., keeping the same video numbers. In my experiment, I firstly sample videos for Box-Siam until it meets the annotation time requirement for Box annotations. Then based on the previously sampled videos, I continue to sample videos to meet the anno. requirement for SoCL-Siam.}}
%\NOTE{you can run the "same video, more frames" experiment for the supplemental. BTW, I remove the previous sentences, so we don't draw attention to it.}

 \begin{figure}
\begin{center}
\vspace{-0.2cm}
   \includegraphics[width=0.75\linewidth]{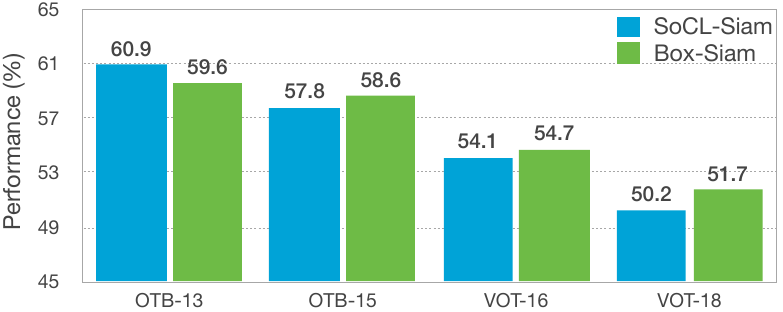} %overall6.eps
\end{center}
\vspace{-0.6cm}
 \caption{Comparison of Box-Siam and SoCL-Siam trained on the whole GOT-10k dataset. The annotation time costs of Box-Siam and SoCL-Siam are respectively 3,967 and 880 hours. The evaluation is conducted  on OTB13/15 (AUC) and VOT16/18 (Accuracy).}
 \vspace{-0.3cm}
\label{same_training_data}
\end{figure}

The results on OTB-13 and OTB-15 are presented in Fig.~\ref{same_anno_otb13} and Table \ref{same_anno_otb15}.
The proposed SoCL-Siam outperforms Box-Siam by large margins for each annotation time cost on OTB-13, indicating that our SoCL effectively learns tracking representations from  low-cost point annotations. 
One interesting phenomena is that the performance gap is larger (56.7 vs.~52.0) for very limited annotation cost  (e.g., 110 hours). This is mainly because: 1) SoCL learns features by comparing objects from the same video and other videos, while Box-Siam treats each video independently, thus Box-Siam is more likely to overfit on small number of training videos; 2) relatively more \jimmyy{video frames} can be used by SoCL due to the less per-frame annotation time cost, even though the total annotation hours are limited.
%
% In Table \ref{same_anno_otb15}, 
 Our SoCL-Siam also achieves better performance than Box-Siam under each total annotation time costs on OTB-15 (Table \ref{same_anno_otb15}).
%  This indicates that our SoCL can effectively learn tracking representations from these cheap point annotations. 

\begin{table}[t]
  \newcommand{\tabincell}[2]
  \centering
  \caption{Comparison of Res18-CF and SoCL-CF.
 % \NOTE{UAV123 has 2 values...AUC and ?}
  % on OTB13/15 (DPR/AUC), UAV123 (AUC), VOT18 (EAO) and LaSOT (AUC). 
%  The best results are highlighted.
  } 
   \vspace{-0.3cm}
  \label{cf_comparison}
 \footnotesize
    \centering
    \begin{tabular}{@{}c@{\hspace{0.2cm}}c@{\hspace{0.2cm}}c@{\hspace{0.2cm}}c@{\hspace{0.2cm}}c@{\hspace{0.2cm}}c@{}}
    \Xhline{\arrayrulewidth}
    \multirow{2}{*}{Method}&  
  OTB13& OTB15 &UAV123 &VOT18&LaSOT \cr
  & DPR/AUC & DPR/AUC & DPR/AUC & EAO & AUC
  \cr   \Xhline{\arrayrulewidth}  
     Res18-CF              &91.3/68.5   &	87.8/65.9 &70.1/51.8 & 0.207 & 31.0     \cr  
     SoCL-CF &\textbf{92.7/69.6}     &\textbf{89.7/67.1} &\textbf{72.3/52.6}   &\textbf{0.216} &\textbf{32.2}   \cr    
   \Xhline{\arrayrulewidth}  
   \end{tabular}  
   \vspace{-0.28cm}
\end{table}

%\vspace{-0.1cm}
We next compare SoCL-Siam and Box-Siam trained on the whole GOT-10k. Note that the total annotation time cost for Box-Siam is much larger than that of  SoCL-Siam (3,967 versus 880 hours).
%
%The comparison is shown in 
Fig.~\ref{same_training_data} shows the performance on various test datasets.
Although SoCL-Siam only uses weak supervision and much less total annotation time,
% of point annotations, 
SoCL-Siam still achieves
%As can be seen, even only using the weakly supervised point annotations and much less total annotation time, SoCL-SiamFC still achieves 
comparable results to Box-Siam, especially on the OTB datasets. 
\abc{Despite the weak supervision of point annotations that do not contain scale information, the proposed SoCL effectively learns discriminative features by exploring the relationships between targets and hard negative samples in each training mini-batch.}

%fully explores relationships between targets and hard negative examples in each mini-batch training, thus learning discriminative features.
% on the GOT-10k dataset with diverse object classes.

\begin{table}[t]
\vspace{-0.2cm}
  \newcommand{\tabincell}[2]
  \centering
  \caption{Comparison of total fees for  BCE learning schema from box annotations and the proposed SoCL schema w/ point annotations on GOT-10k. Amounts are in US dollars.}  
   \vspace{-0.2cm}
  \label{fee_cost}
\footnotesize
    \centering
    \begin{tabular}{@{}cccc@{}}
    \Xhline{\arrayrulewidth}
    \multirow{1}{*}{Schema}& \multicolumn{1}{c}{Anno. Fee}&\multicolumn{1}{c}{Training Fee}&\multicolumn{1}{c}{Total} \cr
     \Xhline{\arrayrulewidth}  
     BCE w/ Boxes                 &\$50,400 &\$7.2 & \$50,407.2      \cr  
     SoCL w/ Points 	  &\$7,000 &\$21.6&\$7,021.6   \cr    
   \Xhline{\arrayrulewidth}  
   \end{tabular}  
   \vspace{-0.35cm}
\end{table}

\begin{table}[t]
\vspace{-0.2cm}
  \newcommand{\tabincell}[2]
  \centering
 %\fontsize{9.2}{8}\selectfont   
  \caption{Comparison of our SoCL-TransT and state-of-the-art deep trackers on GOT-10k \cite{got10k}, TrackingNet \cite{trackingnet} and LaSOT \cite{lasot}. ATC denotes the annotation time cost (hours) of the training set for the tracker.} %overall training frames used to train a tracker.}  
   \vspace{-0.2cm} 
   \label{sotacompare}
   \centering
      \resizebox{\linewidth}{!}{
%\footnotesize
    \begin{tabular}{@{}c@{\hspace{0.1cm}}|@{\hspace{0.2cm}}c@{\hspace{0.1cm}}|@{\hspace{0.1cm}}c@{\hspace{0.1cm}}c@{\hspace{0.05cm}}|@{\hspace{0.15cm}}c@{\hspace{0.1cm}}c@{\hspace{0.1cm}}|@{\hspace{0.15cm}}c@{\hspace{0.15cm}}c@{\hspace{0.1cm}}c@{\hspace{0.05cm}}c} 
    \Xhline{\arrayrulewidth}
    \multirow{2}{*}{Trackers}&\multirow{2}{*}{ATC}& \multicolumn{2}{c}{GOT-10k} & \multicolumn{2}{c}{TrackingNet}& \multicolumn{2}{c}{LaSOT} \cr
    %\cmidrule(r){6-7} \cmidrule(r){8-10} \cmidrule(r){11-13}  
     &		&AO&SR$_{0.5}$&AUC&P$_{Norm}$&AUC&P& \cr   \Xhline{\arrayrulewidth}   
     
     %TransT \cite{transt}   	 &11.4M     		&1.4K                    		                       &72.3&82.4       &81.4                  &86.7&64.9&69.0\cr 
     TransT-GOT \cite{transt}   		&4.0K                    		                       &67.1&76.8       &-           &-&-&-\cr
     KYS \cite{kys}   	 	      		&11.9K                               &63.6&75.1         &74.0&80.0	&55.4&-\cr      
     Ocean  \cite{ocean}        &34.6K 	                  &61.1&   72.1         &-				&-&56.0&56.5\cr
    SiamFC++ \cite{siamfc++}	&42.5K                               &59.5&69.5       &75.4                   &80.0&54.4&54.7\cr  
    SiamRPN++ \cite{SiamRPN_plus}                   &30.6K                               &51.7&61.6     &73.3&80.0                   &49.6&49.1\cr          
    DiMP \cite{dimp}   	 	    		&19.0K                     		                       &61.1&71.7       &74.0&80.1                   &56.9&56.7\cr  
    ATOM \cite{atom}   	 	    		&15.0K                     		                       &55.6&63.4       &70.3&77.1                   &51.5&50.5\cr
    ROAM++ \cite{roam}   	 	    		&26.6K                     		                       &46.5&53.2       &67.0&75.4                   &44.7&44.5\cr
    CGACD \cite{cgacd}                             &34.6K       &-&-  &71.1&80.0 &51.8&- \cr
     D3S \cite{cyclesiam}   	 	 		&4.3K                     		                       &59.7&67.6       &72.8&76.8                  &-&-\cr  
     %CGACD \cite{cgacd}                             &34.6K       &-&-  &71.1&80.0 &51.8&
   % CGACD \cite{cyclesiam}   	 	&2020    		&1.2K                     		                       &-&-       &0.735&0.191                   &0.377&0.750\cr  
  
    \hline
    \textbf{SoCL-TransT}                 	  		&\textbf{1.2K}                    			            &62.2&72.4       &75.0&80.5                  &56.0&56.9\cr
   \Xhline{\arrayrulewidth}  
   \end{tabular}  
   }
   \vspace{-0.55cm}
\end{table}

\subsection{Improving correlation-filter trackers}
\abc{We next show that our SoCL also improves online correlation filter trackers by learning better feature representations.}
%, specifically correlation filters.
We show the comparison between the proposed SoCL-CF and the baseline Res18-CF in Table \ref{cf_comparison}.  %Note that Res18-CF adopts the ImageNet pre-trained ResNet-18 as the feature extractor. 
SoCL-CF achieves consistent improvements over Res18-CF on all the five test datasets. This demonstrates that the proposed SoCL are beneficial for both offline-learning Siamese and online-learning CF trackers. In addition, the total annotation time cost (880h) for gaining these improvements is  acceptable.

% compared with the bounding box cost (3,967h).

% This also implies that 88.8 annotation hours for each person when 10 persons are employed for annotating, and the whole annotating task can be finished in several days. 

%\NOTE{I'm thinking that point-supervision could be a low-cost way to do domain adaptation of the CF tracker. It would be interesting to run the CF experiment using different annotation time cost, to see how many annotations are actually needed to get saturated performance.} \jimmyy{Yes, we can put it in the supplementary. Training w/ pre-trained res-18 for CF is very efficient, so we can also do this experiment.}

\vspace{-0.1cm}
\subsection{Comparison of Total Fees}
We compare the total fee (dollar cost) between the two learning schemas: our SoCL using point annotations  (SoCL w/ Points) and traditional BCE  using bounding box annotations (BCE w/ Boxes) \cite{SiamFC,siamdw}.
 Each bounding box annotation costs \$0.036 using Amazon Mechanical Turk, while each center point annotation costs \$0.005 \cite{clickpoint}.
 Since GOT-10k contains about 1.4M instances, the overall annotation fees  are \$50,400 and \$7,000 for BCE and SoCL. For training time, SiamFC takes about 8 hours for training on GOT-10k with a single GPU card, while ours SoCL takes about 24 hours. 
 A single V100 GPU (p2.xlarge) instance on Amazon EC2 costs 
 %about 
% We use the Amazon EC2 Price to compare the training fee cost. A p2.xlarge (i.e., with a single V100 GPU) instance takes about 
\$0.90 per hour, thus yielding training costs of \$7.2 and \$21.6 for BCE and SOCL. As shown in Table \ref{fee_cost}, 
\abcn{despite having longer training times, our  SoCL w/ Points schema is over 7$\times$ less expensive than BCE w/ Boxes.}

%is significantly less than the BCE-Box schema.
 \vspace{-0.1cm}
\subsection{Comparison with State-of-the-art Trackers}
 %\vspace{-0.1cm}
\jimmyy{We compare our SoCL-TransT %(described in Sec. \ref{ta}) 
with state-of-the-art deep trackers on GOT-10k \cite{got10k}, TrackingNet \cite{trackingnet} and LaSOT \cite{lasot} in Table \ref{sotacompare}.
% shows the comparison.
 Our SoCL-TransT %learned with the proposed new annotation schema
  only requires 1.2K annotation hours on GOT-10k, which is significantly lower than those of  other trackers. 
  Compared to the fully supervised baseline TransT-GOT, 
  our SoCL-TransT achieves 92.7\% and 94.3\% of the AO and SR$_{0.5}$  performances of the fully-supervised baseline, while reducing annotation cost by 70\%.
%  TransT-GOT is our baseline, which is trained on the GOT-10k dataset with manually labeled bounding boxes. The comparison on GOT-10k shows that our SoCL-TransT can respectively achieve 92.7\% and 94.3\% performance of TransT-GOT in terms of AO and SR$_{0.5}$ metrics, and meanwhile reducing annotation cost by 70\%.
 In addition, SoCL-TransT performs favorably against state-of-the-art deep trackers on LaSOT and TrackingNet, even thought it hass low annotation cost.}

%\NOTE{"We believe that 1) the gap could be closed by increasing the overall annotation time cost of SoCL-TransT and 2) SoCL-TransT  is more probably to achieve better performance than TransT-GOT under the same annotation cost." -- this is speculation, you can try these experiments for the supplemental/rebuttal}

\begin{table}[t]
\vspace{-0.2cm}
  \newcommand{\tabincell}[2]
  \centering
  \caption{Comparison of SoCL trained on a subset of GOT-10k w/ and w/o added annotation noise.}
  %The evaluation is conducted on OTB15 (AUC), VOT16 (Accuracy), VOT18 (Accuracy), UAV123 (AUC) and GOT-10k validation (AO).}  
   \vspace{-0.3cm}
  \label{noisy_comparison}
 \footnotesize
    \centering
    \begin{tabular}{@{}c@{\hspace{0.15cm}}c@{\hspace{0.15cm}}c@{\hspace{0.15cm}}c@{\hspace{0.15cm}}c@{\hspace{0.15cm}}c@{}}
    \Xhline{\arrayrulewidth}
    \multirow{2}{*}{Method}&  
 OTB15 & VOT16& VOT18&UAV123 & GOT-10k val \cr
 & AUC & Acc & Acc & AUC & AO
 \cr
     \Xhline{\arrayrulewidth}  
     SoCL              &53.8 &51.5	&46.8	&45.0	&47.2      \cr  
      SoCL w/ added noise   &53.6  &51.0 &46.2	&44.2	&46.6   \cr    
      \Xhline{\arrayrulewidth}
     Difference    &0.2  &0.5 &0.6	&0.8	&0.6   \cr    
   \Xhline{\arrayrulewidth}  
   \end{tabular}  
   \vspace{-0.35cm}
\end{table}

\vspace{-0.35cm}
\subsection{Learning from Noisy Point Annotations}
 \vspace{-0.1cm}
In practice,
the employed annotators  may not be well trained or careful enough during the annotation process, which may lead to noisy point annotations.
% with some spatial annotation noise. 
In \cite{clickpoint}, the average error of point annotations, i.e., the distance between the annotated location and the GT,  was found to be 19.5 pixels for object detection tasks.
% has been studied for the object detection task, which is about 19.5 pixels, i.e., the distance between the annotated location and the GT. 
In order to mimic such noisy annotations, we randomly pick 1000 videos from GOT-10K as our dataset, and add a \abcnn{20 pixel shift}
% (20 pixels) 
with random direction to each point annotation. 
% in each video frame. 
We use this noisy dataset to train SoCL-Siam, and compare it with the model trained using the dataset without adding noise. 

%The obtained models are further integrated to SiamFC for comparison, which is shown in 

Table \ref{noisy_comparison} shows the results. The performance differences   between SoCL with and without annotation noise %and the noisy SoCL on the evaluated five datasets 
is not large, with drops ranging from 0.2\% to 0.8\%. This indicates that our SoCL can learn robust representations from noisy annotations. 
Specifically, 
%One  reason is that 
SoCL uses soft representations instead of performing strict pixel-wise matching (e.g., binary-cross entropy loss) like previous methods \cite{SiamFC,siamrpn,SiamRPN_plus}, which enables SoCL to be \jimmyy{more} robust to noisy data. 

Furthermore, the EdgeBox proposal generation also makes SoCL robust to annotation noise.
%with our design (see Sec. \ref{datapre}) can also reduce the annotation noise. 
We calculate the average error distance between the GT target center and the mean of the generated proposals' centers in each frame.
The average error distance is 14.1 pixels, which is less than the original 20 pixels of the annotation noise.
%we find that the mean annotation noise in each frame is reduced from 20 pixels (i.e., the noise should be 20 pixels if we only use the random proposal generation) to \textbf{14.1} pixels. 
Thus the generated TOP maps are also robust to noisy annotations. 

 \vspace{-0.3cm}
\section{Conclusion}
 \vspace{-0.1cm}
This paper proposes a novel soft contrastive learning (SoCL) framework to learn tracking representations from low-cost single point annotations. 
To facilitate the learning,  we propose several memory-efficient sample generation strategies including the generation of global and local soft templates and  soft negative samples.
%, and local soft templates. 
Although a large number of samples can be included in SoCL for one mini-batch training, the whole training is memory-efficient and can be conducted on a single GPU (e.g., RTX-3090). We successfully apply the learned representations of SoCL to both Siamese and correlation filter tracking frameworks. Moreover, we design a new framework to train bounding box regression-based trackers. 

\section{Acknowledgment}
 \vspace{-0.1cm}
This research was funded by a Strategic Research Grant (Project No. 7005665) from City University of Hong Kong.

\clearpage

% CVPR 2024 Paper Template; see https://github.com/cvpr-org/author-kit

%\documentclass[10pt,twocolumn,letterpaper]{article}

%%%%%%%%% PAPER TYPE  - PLEASE UPDATE FOR FINAL VERSION
%\usepackage{cvpr}              % To produce the CAMERA-READY version
%\usepackage[review]{cvpr}      % To produce the REVIEW version
% \usepackage[pagenumbers]{cvpr} % To force page numbers, e.g. for an arXiv version

% Import additional packages in the preamble file, before hyperref
%\input{preamble}

% It is strongly recommended to use hyperref, especially for the review version.
% hyperref with option pagebackref eases the reviewers' job.
% Please disable hyperref *only* if you encounter grave issues, 
% e.g. with the file validation for the camera-ready version.
%
% If you comment hyperref and then uncomment it, you should delete *.aux before re-running LaTeX.
% (Or just hit 'q' on the first LaTeX run, let it finish, and you should be clear).
\definecolor{cvprblue}{rgb}{0.21,0.49,0.74}
%\usepackage[pagebackref,breaklinks,colorlinks,citecolor=cvprblue]{hyperref}

%\usepackage{times}
%\usepackage{epsfig}
%\usepackage{graphicx}
%\usepackage{amsmath}
%\usepackage{amssymb}
%\usepackage[ruled]{algorithm2e}
%\usepackage{multirow}
%\usepackage{makecell}
%\usepackage{booktabs}
%\usepackage{amssymb}
%\usepackage{pifont}
%\usepackage{ulem}

%\usepackage[font=footnotesize]{caption}

%\usepackage{paralist}

%\usepackage[dvipsnames]{xcolor}

%\usepackage[pagebackref,breaklinks,colorlinks]{hyperref}

% Support for easy cross-referencing
%\usepackage[capitalize]{cleveref}
\crefname{section}{Sec.}{Secs.}
\Crefname{section}{Section}{Sections}
\Crefname{table}{Table}{Tables}
\crefname{table}{Tab.}{Tabs.}

\section{Supplementary Material}

In this supplementary material, we provide additional quantitative comparisons, qualitative visualizations and ablation studies. Section \ref{A} provides the comparison between our SoCL-TransT and TransT \cite{transt} under the same annotation time cost. Section \ref{B} shows the comparison between our SoCL-Siam and Box-Siam with the same annotation time cost and same training videos. Section \ref{C} contains the ablation study on the usage of projection head in both Siamese and correlation filter trackers. We then introduce the detailed preprocessing step  in Section \ref{D} in order to obtain smoother target objectness prior (TOP) maps for more effective representation learning. Finally, Section \ref{E} shows the qualitative visualization of the soft sample generation, including both global soft template (GST) and soft negative sample (SNS) generation.

\renewcommand\thesection{\Alph{section}}

\section{Comparison with Same Annotation Time Cost using TransT} \label{A}
In this section, we compare the proposed SoCL-TransT to its fully supervised baseline (i.e.,TransT \cite{transt}) trained with bounding boxes under the same time cost of annotation. Note that SoCL-TransT is trained on the whole GOT-10k dataset with point annotations, and its total annotation time cost is about 1.2K hours. %For fair comparison, 
We randomly sample training videos with bounding box annotations from GOT-10k to meet the same annotation time requirement (1.2K hours), and then use these videos to train TransT. As illustrated in Table \ref{sotacompare}, our SoCL-TransT achieves better performance than TransT in terms of all the metrics on the three large-scale tracking datasets. For example, SoCL-TransT achieves  favorable AUC, P$_{Norm}$ and P on the LaSOT dataset by respectively improving 7.1\%, 8.9\% and 10.1\% compared to TransT, which demonstrates the effectiveness of the proposed new annotation schema for training scale regression-based trackers.

\begin{table}[tbp]
%\vspace{-0.1cm} 
  \newcommand{\tabincell}[2]
  \centering
 %\fontsize{9.2}{8}\selectfont   
  \caption{Comparison of SoCL-TransT and TransT trained using the same annotation time cost (i.e., 1.2K hours) on GOT-10k \cite{got10k}, TrackingNet \cite{trackingnet} and LaSOT \cite{lasot}. The best results are highlighted.} %overall training frames used to train a tracker.}  
  % \vspace{-0.3cm} 

   \label{sotacompare}
   \centering
      \resizebox{\linewidth}{!}{
%\footnotesize
    \begin{tabular}{@{}c@{\hspace{0.1cm}}|@{\hspace{0.2cm}}c@{\hspace{0.1cm}}@{\hspace{0.1cm}}c@{\hspace{0.1cm}}c@{\hspace{0.15cm}}|@{\hspace{0.15cm}}c@{\hspace{0.1cm}}c@{\hspace{0.1cm}}c@{\hspace{0.1cm}}|@{\hspace{0.2cm}}c@{\hspace{0.15cm}}c@{\hspace{0.15cm}}c} 
    \Xhline{\arrayrulewidth}
    \multirow{2}{*}{Trackers}& \multicolumn{3}{c}{GOT-10k} & \multicolumn{3}{c}{TrackingNet}& \multicolumn{3}{c}{LaSOT} \cr
    %\cmidrule(r){6-7} \cmidrule(r){8-10} \cmidrule(r){11-13}  
     &		AO&SR$_{0.5}$&SR$_{0.75}$&AUC&P$_{Norm}$&P&AUC&P$_{Norm}$&P \cr   \Xhline{\arrayrulewidth}   
     
     %TransT \cite{transt}   	 &11.4M     		&1.4K                    		                       &72.3&82.4       &81.4                  &86.7&64.9&69.0\cr 
     TransT \cite{transt}   		&59.1                    		                       &68.0&51.9       &72.4           &76.4&67.1&48.9&50.1&46.8\cr
          %CGACD \cite{cgacd}                             &34.6K       &-&-  &71.1&80.0 &51.8&
   % CGACD \cite{cyclesiam}   	 	&2020    		&1.2K                     		                       &-&-       &0.735&0.191                   &0.377&0.750\cr  
  
    \hline
    \textbf{SoCL-TransT}                 	  		&\textbf{62.2}                   			            &\textbf{72.4}&\textbf{52.5}       &\textbf{75.0}&\textbf{80.5}                  &\textbf{71.1}&\textbf{56.0}&\textbf{59.0}&\textbf{56.9}\cr
   \Xhline{\arrayrulewidth}  
   \end{tabular}  
   }
\end{table}

\section{Comparison with Same Annotation Time Cost and Same Training Videos} \label{B}
%\NOTE{The experiment is running: we only need to retrain Box-Siam, so this experiment is easy to do.}
In Section 4.3 of the main paper, under the same annotation time cost, the training sets for baselines or SoCL are selected by sampling whole videos to ensure the same time cost. However, this means that the baseline methods are trained on fewer  videos (possibly seeing less backgrounds and  less objects) compared to SoCL.  
%
%SoCL-Siam can use more sampled videos with point annotations for training due to the low point annotation time cost.
%
 In this section, we guarantee that SoCL-Siam and Box-Siam use the same number of training videos and same annotation cost. Specifically, for each video, SoCL-Siam uses all its video frames while Box-Siam uses about 22.2\% (i.e., 1/4.5) of the frames.
 %to meet the same annotation time cost with Box-Siam. 
 The comparison is shown in Table \ref{same_anno_video_otb13}. We can see that our SoCL-Siam still significantly outperforms Box-Siam under various annotation time costs, which shows the superiority of our SoCL and demonstrates that SoCL can also learn effective temporal correspondences from soft representations for visual tracking.

\begin{table}[t]
  \newcommand{\tabincell}[2]
  \centering
  \caption{Comparison of Box-Siam and SoCL-Siam trained using the same annotation time costs (i.e., hours) and the same number of training videos in terms of AUC on OTB-13. The best results are highlighted.}  
   %\vspace{-0.2cm}
  \label{same_anno_video_otb13}
%\footnotesize
    \centering
   \begin{tabular}{ccccc}
    \Xhline{\arrayrulewidth}
    \multirow{1}{*}{Annotation time cost}&  
  \multicolumn{1}{c}{110h}& \multicolumn{1}{c}{220h}&\multicolumn{1}{c}{440h}&\multicolumn{1}{c}{880h} \cr
     \Xhline{\arrayrulewidth}  
     Box-Siam              &52.2   &54.3 &58.1 & 58.8      \cr  
     SoCL-Siam &\textbf{56.7}	  &\textbf{58.2} &\textbf{59.8}   &\textbf{60.9}   \cr    
   \Xhline{\arrayrulewidth}  
   \end{tabular}  
\end{table}

\section{Ablation Study on Projection Head}  \label{C}

\begin{table}[tbp]
  \newcommand{\tabincell}[2]
  \centering
  \caption{Ablation study on projection heads: AUCs obtained by using different trackers w/ and w/o projection heads on OTB-13. The best results are highlighted.}  
   %\vspace{-0.2cm}
  \label{proj}
%\footnotesize
    \centering
   \begin{tabular}{ccccc}
    \Xhline{\arrayrulewidth}
    \multirow{1}{*}{}&  
  \multicolumn{1}{c}{w/ proj. head}& \multicolumn{1}{c}{w/o proj. head}\cr
     \Xhline{\arrayrulewidth}  
     SoCL-Siam &	49.5  &\textbf{60.9}   \cr    
     SoCL-CF &\textbf{69.6}	  & 68.8    \cr    
   \Xhline{\arrayrulewidth}  
   \end{tabular}  
\end{table}

 \begin{figure*}
\begin{center}
   \includegraphics[width=0.7\linewidth]{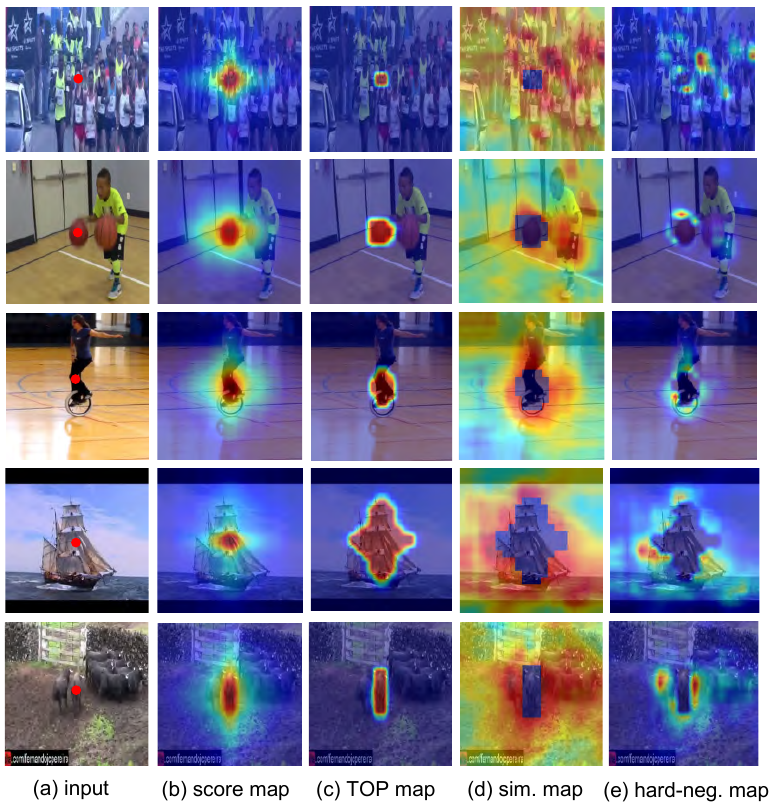} %overall6.eps
\end{center}
\vspace{-0.5cm}
 \caption{Qualitative visualization of global soft template (GST) and soft negative sample (SNS) generation in the proposed SoCL framework. (a) Input search images that contain both target and background regions. 
% \NOTE{add the point annotation to the input images.}
 (b) Score maps generated via proposal generation and objectness measurement \cite{obj}. (c) TOP maps generated by applying the softmax function on the score maps.
 (d) the similarity map with target responses masked out using the background selection function.
 (e) applying the softmax to (d) yields the hard negative map, which is used to select features for the SNS.
%  (d) \NOTE{add this column to the figure} the similarity map based on cross-correlation between the generated GST and the search feature map.
 % (e) the similarity map with target responses masked out using the background selection function.
%  We generate GSTs via the guidance of TOP maps, and use the GSTs to perform cross-correlation on the search feature maps in order to obtain similarity maps. The high responses for the target in the similarity maps are masked out using the TOP maps and the background selection function. 
%(f) applying the softmax to (e) yields the hard negative map, which is used to select features for the SNS.
%  (e) We generate hard-negative maps by applying the softmax function on the similarity maps.  The SNSs are generated as a weighted sum of the search feature maps over spatial locations, weighted by the hard-negative maps.
 }
\label{qualitative}
\end{figure*}

The usage of a projection head has been well explored in the contrastive learning community \cite{simclr,simclrextend}. Commonly, using a projection head for end-to-end contrastive learning can learn better feature representations for some typical downstream tasks, e.g., image classification. However, there is no empirical study to explore its usage on visual tracking, i.e., whether it is beneficial for learning robust tracking representations. Note that the projection head we use is implemented as a three-layer perceptrons with a single hidden layer of $K$ units, where $K$ is set to the dimension of features extracted from the backbone. The output of the final perceptron is a 64-dimensional vector. 

We conduct this ablation study on two different types of tracking frameworks:  offline learning-based Siamese and online learning-based correlation filter (CF) trackers. Specifically, we put the projection head after the backbone networks used in SoCL-Siam and SoCL-CF\footnote{For the ResNet-18 backbone used in SoCL-CF, we remove its average pooling and fully-connected layers, and modify its stride in {\em{Layer4}} to 1, so that the final output in the
feature space can have a relatively large spatial size, which is more beneficial for soft sample generation.} to further extract features for contrastive learning.

The results are presented in Table \ref{proj}. SoCL-Siam with the projection head degrades the performance, while SoCL-CF with the projection head achieves better performance than its variant without a projection head.
%using it.
 The main reason is that SoCL-Siam directly uses the extracted backbone features for online tracking without further updating. The learning of SoCL-Siam without the projection head is consistent with its online tracking process, thus leading to better performance. Moreover, SoCL-Siam with the projection head treats the backbone network as the intermediate layers, which facilitates the backbone to learn to encode more detailed and rich information into features. These features are not good for offline learning-based trackers without further online updating. Compared with SoCL-Siam, SoCL-CF can benefit from these features due to its powerful online updating mechanism.

\section{Preprocessing of TOP Map}  \label{D}

\begin{table}[tbp]
  \newcommand{\tabincell}[2]
  \centering
  \caption{AUCs obtained by using various $\eta$ on OTB-13. The best results are highlighted.}  
   %\vspace{-0.2cm}
  \label{abeta}
%\footnotesize
    \centering
   \begin{tabular}{ccccc}
    \Xhline{\arrayrulewidth}
    \multirow{1}{*}{$\eta$}&  
  \multicolumn{1}{c}{5\%}& \multicolumn{1}{c}{10\%}&\multicolumn{1}{c}{15\%}&\multicolumn{1}{c}{20\%} \cr
     \Xhline{\arrayrulewidth}  
     SoCL-Siam &{60.0}	  &\textbf{60.9} &{59.5}   &{59.1}   \cr    
   \Xhline{\arrayrulewidth}  
   \end{tabular}  
\end{table}

The target objectness prior (TOP) maps are generated by applying the softmax function on the score maps, which are calculated via the generated proposals (see Sec. 3.1 of the main paper). In practical implementation, we find that the TOP maps may 
 have extremely large peaks on the annotated locations, which makes the generation of GSTs excessively focus on these locations. This is because there are large peaks in the score maps, and the softmax function assigns too much weights on these locations. To alleviate this problem, we use a simple max clip operation to clip maximum values in score maps. Specifically, given a score map, we firstly set an adaptive clip threshold $\eta$. Then we calculate the mean score of the top-$\eta$ scores in the score map. The calculated mean score is used to perform the max clip in  the score map, so that the scores with large values will have the same value, and thus more locations will be selected by applying the softmax function.
 
Table \ref{abeta} shows the performance obtained by using various $\eta$ for the clip threshold.  The optimal performance is achieved by setting $\eta=10\%$. Setting $\eta$ to larger values (e.g., 15\% and 20\%) may cause the generated TOP maps to excessively focus on background regions, thus degrading the performance.

\section{Qualitative Visualization} \label{E}
Fig.~\ref{qualitative} shows the qualitative visualization of GST and SNS  generation. The generation of SNSs tends to aggregate features from discriminative regions, e.g., target boundary regions (see the third and fourth rows) and hard negative counterparts (see the first, second and fifth rows). Note that all the maps in Fig.~\ref{qualitative} are interpolated to the input image size for visualization.

{
    \small
    \bibliographystyle{ieeenat_fullname}
    \bibliography{main}
}

\end{document}